\documentclass[10pt,twocolumn,letterpaper]{article}

\usepackage{wacv}              % To produce the CAMERA-READY version
\usepackage[utf8]{inputenc}
\usepackage{url}
\usepackage{threeparttable}

\definecolor{wacvblue}{rgb}{0.21,0.49,0.74}
\usepackage[pagebackref,breaklinks,colorlinks,allcolors=wacvblue]{hyperref}
\usepackage{float}
\def\wacvPaperID{2584} % *** Enter the WACV Paper ID here
\def\confName{WACV}
\def\confYear{2027}

\title{PanoDent-KD: From Generalist Knowledge Distillation to Cross-Dataset Specialist Detection and Embedded Deployment in Panoramic Radiographs}

\author{
    Muhammad Waqas Ashraf~\textsuperscript{1,\dag} \quad
    Imad Ali Shah~\textsuperscript{2} \quad
    Ali Hassan~\textsuperscript{1}
}

\begin{document}
\maketitle

% Footnotes at the bottom of the first page
\begingroup
\renewcommand{\thefootnote}{%
    \ifcase\value{footnote}\or 1\or 2\or \dag\or *\fi}

\footnotetext[1]{~National University of Sciences and Technology, Pakistan}
\footnotetext[2]{~University of Galway, Ireland}
\footnotetext[3]{~Corresponding author: \texttt{waqas51006@gmail.com}}
% \footnotetext[4]{~Contributed equally}
\endgroup

% \begingroup
% \renewcommand{\thefootnote}{%
%     \ifcase\value{footnote}\or *}

% \footnotetext[1]{Correspondence: \texttt{waqas51006@gmail.com}}
% \endgroup

\begin{abstract}
Reliable panoramic-radiograph detection requires consistency across datasets with different class definitions and deployment software. We present PanoDent-KD, an auditable 1280-pixel YOLO26m pipeline linking a 12-class source generalist, feature-level knowledge distillation (KD), exact classifier-row transfer, teacher-free five-class specialization, and embedded deployment. A common training and evaluation pipeline compares three source-training arms across three seeds: a supervised baseline (B1), single-teacher KD (STKD), and multi-teacher KD (MTKD). Five shared classifier rows initialize nine OralXrays-9 specialists, while the backbone, FPN–PAN neck, and dual-detection-head topology are retained unchanged. On the 2,686-image test subset, all three arms achieve macro box F1 near 0.82. B1 achieves the highest three-seed arm-level mean macro box F1 (0.8239$\pm$0.0027) and $mAP_{50:95}$ (0.6130$\pm$0.0067). STKD seed 1, selected after development but before final testing, was used for deployment. Its Jetson Orin Nano Super FP16 TensorRT engine achieves 77.03-ms median and 80.86-ms P95 prediction-call latency,  throughput of 12.90 images/s, and gross total-module energy of 1.39 J/image, with high numerical agreement between within-device PyTorch and TensorRT FP16 inference. These findings demonstrate the feasibility of an auditable, device-specific deployment workflow, but do not establish clinical validity or improved target-domain accuracy from KD.
\end{abstract}
    
\section{Introduction}
Panoramic radiographs summarize the dentition, jaws, restorations, and adjacent structures, but their breadth creates scale variation, superposition, artifacts, and protocol-dependent boundaries. Hospital-scale resources now support multi-object study: OralXrays-9 contains 12,688 radiographs and nine anomaly classes~\cite{cite7}; UMFIH provides development and external cohorts~\cite{cite24, cite25}; and a related 12-class task removes two exceptionally rare labels~\cite{cite11}. DENTEX~\cite{cite13}, Tufts~\cite{cite28}, and CariesXrays~\cite{cite6} illustrate the diversity of acquisition and ontology.

The main challenge is transferring models between datasets, not only model size. Tooth numbering~\cite{cite40}, apical-lesion~\cite{cite12}, bone-loss~\cite{cite19}, and restoration systems~\cite{cite1} learn recognizable structures, yet internal accuracy does not establish robustness to another hospital or prevalence. Dental and medical-imaging guidance therefore emphasizes intended use, provenance, partitioning, reference standards, uncertainty, and deployment context~\cite{cite15, cite29, cite35, cite38, cite41, cite44}. Radiographic models can deteriorate between institutions~\cite{cite47}, exploit shortcuts~\cite{cite8}, and hide important failure strata behind aggregates~\cite{cite27}.
Distillation is an appealing bridge: a large teacher may organize features for a compact student. But a small source validation gain may arise from contamination, seed variation, metric policy, or checkpoint choice and disappear after target fine-tuning. Leakage analyses motivate exact-content auditing~\cite{cite18, cite37}, while domain theory and dataset-bias studies predict that source risk does not determine target risk~\cite{cite2, cite39}.

PanoDent-KD investigates whether source feature distillation survives external source testing, five-class specialization, and hardware-specific export. We audit source partitions, harmonize evaluation, train B1/STKD/MTKD over three seeds, and reserve the final test until after model selection. All nine source checkpoints independently pass transfer preflight; five shared classifier rows initialize teacher-free OralXrays-9 specialists. One deployment checkpoint is frozen before final-test result access and exported as a pure student.

\begin{figure*}[!t]
    \centering
    \includegraphics[width=0.98\textwidth]{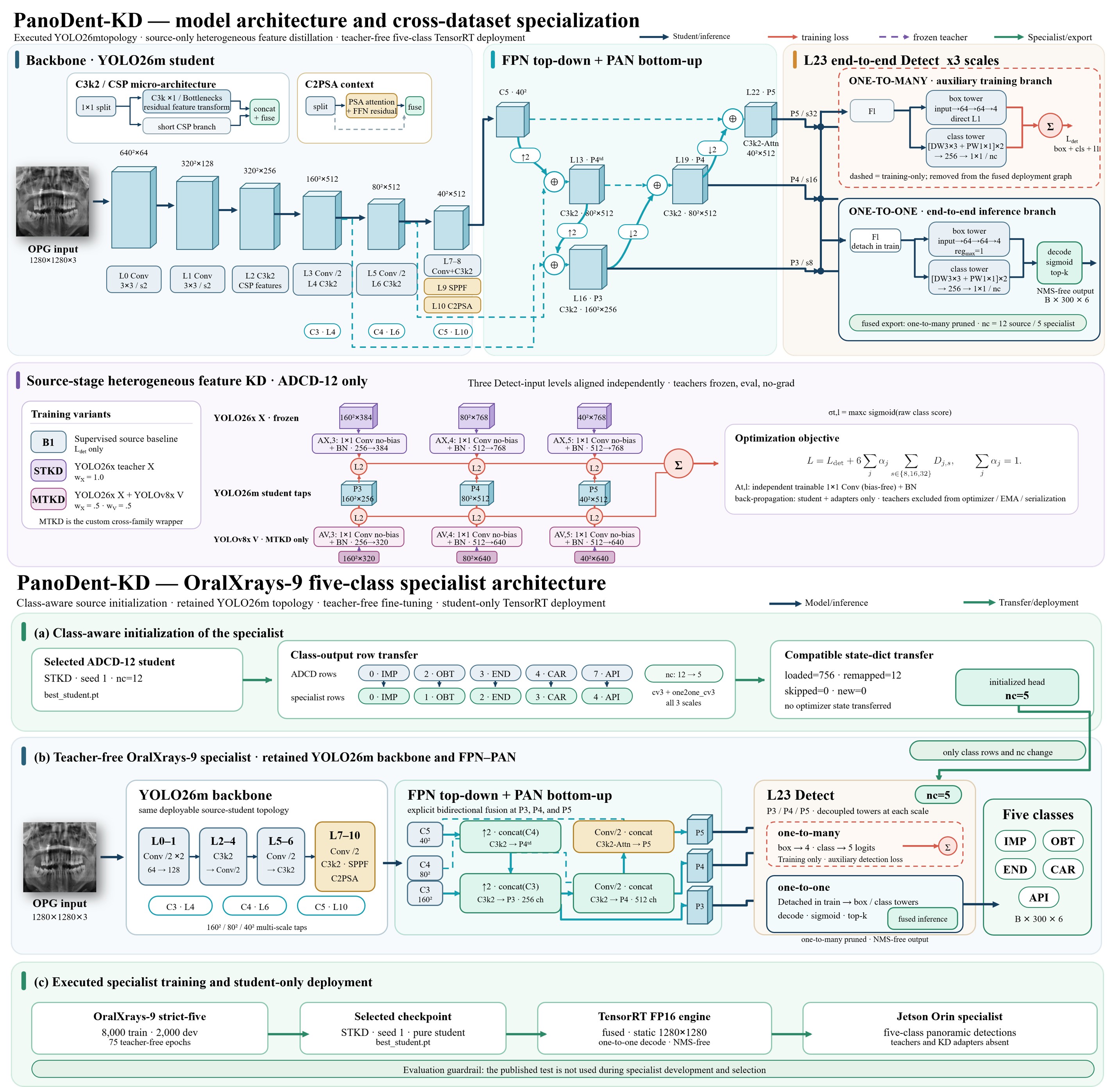}
    \vspace{-1ex}
    \caption{PanoDent-KD architecture. (top) YOLO26m source student ($1280\times1280$); STKD/MTKD add frozen teachers via $1\times1$ Conv+BN adapters. (bottom) The specialist keeps this topology, maps source rows $[0,2,3,4,7]$ to target rows $[0,1,2,3,4]$ (756 tensors loaded, 12 remapped), trains 75 teacher-free epochs, and exports the pure student in FP16.}
    \label{fig:architecture}
    \vspace{-2ex}
\end{figure*}

The audit bounds every claim. Forty-two exact-image groups cross source train and validation sets, so the released split supports historical comparison rather than a new state-of-the-art claim. Target train and development have no exact-content, stem, or inode overlap. A blinded review found two test images that also appear in training: the frozen exclusion rule creates the 2,686-image test subset. As panoramic patterns can remain identifying~\cite{cite9}, the published panel is hash-selected, metadata-stripped, manually screened, and used under the retained non-commercial publication terms.

\begin{figure*}[t]
    \centering
    \includegraphics[width=\textwidth]{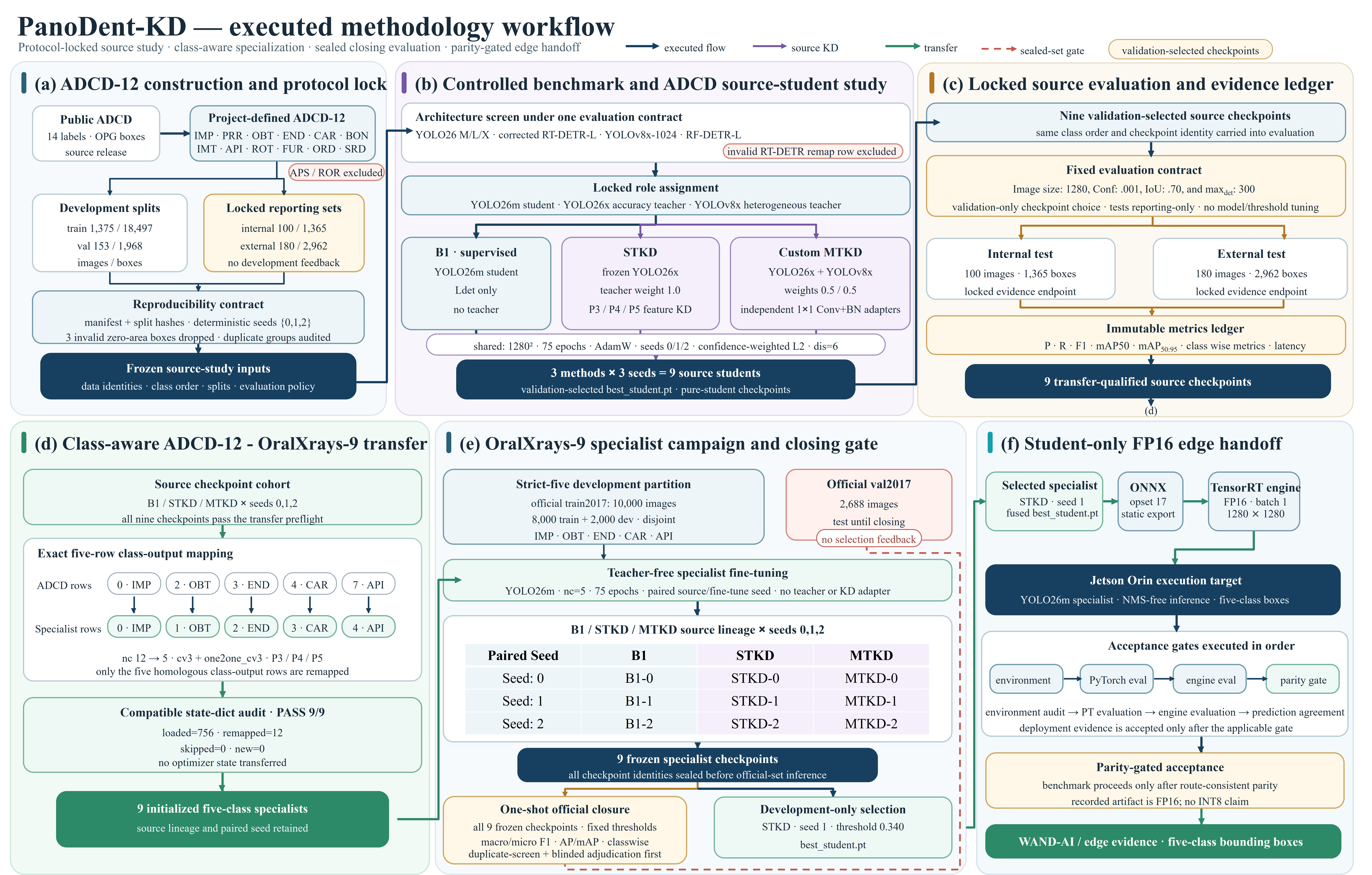}
    \caption{PanoDent-KD workflow. (a) Data plan; (b) source architecture screen and B1/STKD/MTKD training; (c) source testing; (d) five-row transfer to OralXrays-9; (e) nine teacher-free specialists, STKD seed 1 selected on development data (confidence 0.340), single final evaluation; (f) FP16 TensorRT export and Orin checks. No teacher or KD adapter is used in specialist training or inference.}
    \label{fig:methodology}
    \vspace{-1ex}
\end{figure*}

Figure~\ref{fig:architecture} separates the executed source student and its OralXrays-9 specialist transformation: the former exposes the deployable backbone–neck–head topology and source-only KD attachments, whereas the latter expands the exact five-row transfer, teacher-free specialization, and student-only edge route. The main contribution is not a new KD method. It is a checked workflow combining (i) three-seed source and nine-run target comparisons, (ii) recorded class-row and weight transfer, (iii) fixed final evaluation, and (iv) runtime-agreement checks and real-device speed, power, and memory measurements.

\section{Related work}
\subsection{One-stage detectors}
One-stage YOLO detectors cast localization and classification as dense prediction~\cite{cite32}. Feature pyramids~\cite{cite20} and path aggregation~\cite{cite22} retain spatial detail while propagating semantics. The executed lineage includes versioned Ultralytics software~\cite{cite16} and YOLO26's end-to-end, NMS-free path~\cite{cite17}. Comparisons still require one evaluator because preprocessing, class loading, confidence filtering, and maximum detections can change rankings independently of learned features.

\subsection{Convolutional and transformer detectors}
Set-prediction detectors provide complementary baselines. DETR~\cite{cite5}, Deformable DETR~\cite{cite50}, DINO~\cite{cite48}, and RT-DETR~\cite{cite49} progressively strengthen assignment, multi-scale sampling, and efficient inference; RF-DETR~\cite{cite33} with realtime inference~\cite{cite51} and EdgeCrafter/ECDet~\cite{cite21} extend the design space. ConvNeXt~\cite{cite23} and DINOv3~\cite{cite36} motivate the frozen exploratory teacher. Our screen is empirical rather than compute-matched.

\subsection{Knowledge distillation (KD)}
Existing studies~\cite{cite14, cite16, cite34, cite46} cover KD, feature matching, attention transfer, and multiple teachers~\cite{cite4, cite42, cite43, cite45}. We reuse these methods, rather than introduce a new KD loss. Frozen YOLO teachers guide P3/P4/P5 through 1×1 Conv+BN adapters and confidence-weighted MSE. MTKD weights the YOLO26x and YOLOv8x teachers equally, where the latter was originally trained at 1024 pixels. Both run at 1280 pixels during distillation. Therefore, we test whether source KD benefits remain after target training.

\begin{table*}[t]
\centering
\caption{Datasets and Evaluation Roles}
\label{tab:datasets_and_roles}
\begin{threeparttable}
\begin{tabular*}{\textwidth}{@{\extracolsep{\fill}} l r r r l l @{}}
\toprule
\textbf{Partition} & \textbf{Classes} & \textbf{Images} & \textbf{Boxes} & \textbf{Role} & \textbf{Integrity Boundary} \\
\midrule
ADCD train       & 12 & 1,375 & 18,497 & source training     & 42 exact groups cross to val \\
ADCD val         & 12 &  153 &  1,968 & model selection     & Duplicates across split \\
ADCD internal    & 12 &  100 &  1,365 & fixed test          & No listed exact-copy pair \\
ADCD external    & 12 &  180 &  2,962 & external report     & No listed exact-copy pair \\
% \addlinespace
Oral train       &  5 & 8,000 & 25,483 & target training     & No exact/stem/inode dev overlap \\
Oral dev         &  5 & 2,000 &  6,370 & model selection     & Patient independence unverified \\
Test subset      &  5 & 2,686 &  9,552 & one-shot (final test) & 2 records removed before final test \\
\bottomrule
\end{tabular*}
\begin{minipage}{\textwidth}
\vspace{1pt}
\begin{flushleft}
\scriptsize ADCD-12 is a 12-class UMFIH view. The 2,686-image test subset removes the two test images that also appear in training from the published 2,688-image OralXrays-9 pool.
\end{flushleft}
\end{minipage}
\end{threeparttable}
\vspace{-2pt}
\end{table*}

\subsection{Evaluation and deployment discipline}
Previous work highlights seed variation~\cite{cite3, cite30} and TensorRT conversion~\cite{cite26}, but export is not benchmarking: device testing still matters~\cite{cite31}. We combine these established methods with recorded class transfer and fixed target evaluation. We measure agreement between inference methods, device speed, and energy use. These results do not establish clinical performance.

\section{Methodology}
Figure~\ref{fig:methodology} fixes the chronology from source construction to edge handoff, while Fig.~\ref{fig:architecture} exposes the source and specialist topologies. The 1280-pixel YOLO26m student uses a Conv/C3k2 backbone, SPPF–C2PSA context, bidirectional FPN–PAN fusion, and P3/P4/P5 outputs at strides $S=\{8,16,32\}$ with shapes 160\textsuperscript{2} $\times$256, 80\textsuperscript{2}$\times$512, and 40\textsuperscript{2}$\times$512. L23 Detect contains one-to-many training and one-to-one inference branches; reg$_\text{max}$=1 gives direct regression, and the fused route is NMS-free~\cite{cite17}.
For teacher $j$, batch $b$, and spatial location $(u,v)$, the
preferred populated one-to-many logits $z_T$ define the spatial
confidence mask
\begin{equation}
w_{j,s}^{(b)}(u,v)
=
\max_{c\in\{1,\ldots,12\}}
\sigma\!\left(
z_{j,s,c}^{(T,b)}(u,v)
\right)
\label{eq:confidence_mask}
\end{equation}
where $c$ indexes the 12 source classes and $\sigma(\cdot)$ denotes
the sigmoid function.

A bias-free $1\times1$ Conv+BN adapter $A_{j,s}$ aligns each scale, and
$\Delta_{j,s,b,k,u,v}$ denotes an entry of the adapted-student minus
teacher residual. The executed discrepancy is
\begin{equation}
D_{j,s}
=
\operatorname{mean}_{b,k,u,v}
\left[
w_{j,s}^{(b)}(u,v)
\Delta_{j,s,b,k,u,v}^{2}
\right]
\label{eq:distillation_discrepancy}
\end{equation}
With $\lambda=6$ and $s\in\{8,16,32\}$, the total loss is
\begin{equation}
L
=
L_{\mathrm{det}}
+
\lambda\sum_j \alpha_j\sum_{s}D_{j,s},
% \qquad
\text{where }\sum_j\alpha_j=1.
\label{eq:total_loss}
\end{equation}
B1 uses L$_\text{det}$; STKD sets $\alpha=1$; MTKD fixes $\alpha=0.5$ per teacher. Common batch-size scaling is applied. Frozen teachers are absent from the optimizer, EMA, and serialized student.

\subsection{Specialist transfer}
Each of the nine source checkpoints loads all shape-compatible backbone, neck, box-tower and class-tower parameters. Source rows [0,2,3,4,7] (IMP, OBT, END, CAR, API) map to target rows [0,1,2,3,4] in both class branches at P3/P4/P5. The audit records 756 loaded and 12 remapped tensors, with no skipped or newly initialized tensors. Optimizer state is discarded, and specialists train 75 epochs without teachers. Lineage names therefore identify initialization, not target-domain KD.

\subsection{Model selection and evaluation}
The original plan selected B1 based on mean development mAP$_\text{50:95}$. After reviewing development results, but before viewing final test results, a dated change selected STKD seed 1 for deployment. Confidence was fixed at 0.340, with macro box F1 measured at IoU 0.50 using exact box matching. This change was not registered in advance.  Coupled source/target seeds (0$\rightarrow$0,1$\rightarrow$1,2$\rightarrow$2) limit conclusions about the effect of KD alone. mAP$_\text{50}$ and mAP$_\text{50:95}$ use the common evaluator. Macro F1 averages five classes and micro F1 pools TP/FP/FN; both are box, not patient, endpoints. Results are mean$\pm$sample SD (n=3). Paired image bootstraps~\cite{cite10} condition on frozen checkpoints and omit retraining uncertainty. Only three selected checkpoints undergo source testing; all nine pass the transfer checks. The final test subset is evaluated after model selection, while device agreement is checked using development data only.
 
\begin{table*}[t]
\centering
\caption{Source Architecture Screen and KD Lineage Results}
\label{tab:kd_lineage_results}
\begin{tabular*}{\textwidth}{@{\extracolsep{\fill}} l l c r r l @{}}
\toprule
\textbf{Study} & \textbf{Model} & \textbf{Val mAP$_\text{50:95}$} & \textbf{Int. mAP$_\text{50:95}$} & \textbf{Ext. mAP$_\text{50:95}$} & \textbf{Qualification} \\
\midrule
Screen & ECDet-X   & --                 & 0.2122 & 0.1488  & 50 epochs \\
Screen & RF-DETR-L & --                 & 0.3144 & 0.2206  & early stop \\
Screen & RT-DETR-L & --                 & 0.3054 & 0.2107  & no remap \\
Screen & YOLO26-M  & --                 & 0.3132 & 0.2286  & 75 epochs \\
Screen & YOLO26-L  & --                 & 0.3236 & 0.2245  & 75 epochs \\
Screen & YOLO26-X  & --                 & 0.3178 & 0.2505  & 75 epochs \\
Screen & YOLOv8x   & --                 & 0.2981 & 0.2285  & 1024/25 epochs \\
\midrule
KD     & B1        & 0.2236\scriptsize{$\pm$0.0065} & 0.3037 & 0.2284 & seed 0 report \\
KD     & STKD      & 0.2244\scriptsize{$\pm$0.0033} & 0.3230 & 0.2215 & seed 0 report \\
KD     & MTKD      & 0.2246\scriptsize{$\pm$0.0040} & 0.3201 & 0.2164 & seed 1 report \\
\bottomrule
\end{tabular*}
\begin{minipage}{\textwidth}
\vspace{4pt}
\begin{flushleft}
\scriptsize KD validation: mean$\pm$sample SD ($n=3$). Test rows are selected, partly unmatched checkpoints; source selection inherits the forty-two group limitation.
\end{flushleft}
\end{minipage}
\end{table*}

\section{Experiments}
\subsection{Data integrity}
Table~\ref{tab:datasets_and_roles} defines every partition and role. The UMFIH source describes 14 categories~\cite{cite24, cite25}; ADCD-12 follows the 12-class derivative in~\cite{cite11} and uses the ontology abbreviations IMP, PRR, OBT, END, CAR, BON, IMT, API, ROT, FUR, ORD, and SRD. Three boxes with zero width are removed. Image-content checks find forty-two groups of identical images across training and validation. We retain the published split for comparison, but these duplicates prevent reliable claims about small KD gains or state-of-the-art accuracy.

OralXrays-9 reports a patient-level 10,000/2,688 train–test partition~\cite{cite7}, but local patient identifiers are unavailable. The 8,000/2,000 target training/development split has no identical images, but patient independence cannot be confirmed. Before viewing final test results, a blind review identified two radiographs also present in training. Removing these test records left 2,686 images. Results on this subset cannot be directly compared with results on the original 2,688-image test set. The hashed release contains 84,133 annotations versus 84,113 reported in~\cite{cite7}; the unresolved 20-instance difference did not affect model or threshold selection.

\subsection{Training and baselines}
Principal source and target YOLO runs use 75 epochs, batch size 4, 1280-pixel inputs, AdamW, initial learning rate 6.25$\times10^{-4}$, weight decay 5$\times10^{-4}$, three warm-up epochs, AMP, and seeds 0–2. Augmentation is deliberately mild: $\pm$3° rotation, 0.05 translation, 0.10 scaling, and 0.5 horizontal flip; mosaic, mixup, cutmix, and copy-paste are disabled. The source teachers are frozen seed-0 checkpoints. The DINOv3/ConvNeXt-L ablation is a bounded 25-epoch seed-0 validation probe whose +0.005 mAP$_\text{50:95}$ advancement gate is not met.

Table~\ref{tab:kd_lineage_results} reports ECDet-X, RF-DETR-L, guarded RT-DETR-L with class remapping disabled, YOLO26-M/L/X, and YOLOv8x after compatible-engine AP recomputation. Budgets remain unequal: ECDet runs 50 epochs, RF-DETR early-stops at epoch 29 of 100, and YOLOv8x uses 25 epochs at 1024 pixels. An older RT-DETR row is excluded because name-based loading mapped the pretrained COCO car row into dental CAR. This is an empirical family screen, not a controlled compute or latency comparison.

\subsection{Deployment protocol}
The amended STKD seed-1 specialist is fused, exported through ONNX opset 17, and compiled on target as static batch-1, 1280-pixel FP16 TensorRT; INT8 is unused. Agreement testing uses the 2,000-image development split and decomposes desktop validator$\rightarrow$predictor, desktop predictor$\rightarrow$Orin PyTorch, and Orin PyTorch$\rightarrow$TensorRT; no target test image is accessed. A separate performance-only amendment repeats one hash-fixed 500-image development sequence in three trials after 50 disjoint warm-ups each, using CUDA-synchronized prediction calls, page-cached files, 100-ms tegrastats, MAXN$_\text{SUPER}$, and fixed clocks. We exclude acquisition, UI/display, network, cold start, and concurrency.

\begin{table*}[t]
\caption{OralXrays-9 Specialist Results on Development and Test Subset}
\label{tab:oralxrays9_means}
\centering
\resizebox{\textwidth}{!}{%
\begin{tabular}{lrrrrrrr}
\toprule
\textbf{Arm} & \textbf{Dev Proxy F1} & \textbf{Dev $\text{mAP}_{50:95}$} & \textbf{Test Macro F1} & \textbf{Test Micro F1} & \textbf{Test mAP$_\text{50}$} & \textbf{Test mAP$_\text{75}$} & \textbf{Test $\text{mAP}_{50:95}$} \\
\midrule
B1   & 0.8410\scriptsize{$\pm$0.0023} & 0.6456\scriptsize{$\pm$0.0044} & 0.8239\scriptsize{$\pm$0.0027} & 0.8449\scriptsize{$\pm$0.0035} & 0.8541\scriptsize{$\pm$0.0014} & 0.6587\scriptsize{$\pm$0.0021} & 0.6130\scriptsize{$\pm$0.0067} \\
STKD & 0.8428\scriptsize{$\pm$0.0064} & 0.6408\scriptsize{$\pm$0.0123} & 0.8206\scriptsize{$\pm$0.0074} & 0.8418\scriptsize{$\pm$0.0086} & 0.8539\scriptsize{$\pm$0.0019} & 0.6613\scriptsize{$\pm$0.0019} & 0.6103\scriptsize{$\pm$0.0094} \\
MTKD & 0.8402\scriptsize{$\pm$0.0021} & 0.6348\scriptsize{$\pm$0.0060} & 0.8192\scriptsize{$\pm$0.0043} & 0.8404\scriptsize{$\pm$0.0041} & 0.8495\scriptsize{$\pm$0.0039} & 0.6589\scriptsize{$\pm$0.0031} & 0.6035\scriptsize{$\pm$0.0078} \\
\bottomrule
\end{tabular}%
}
\vskip 4pt
\begin{minipage}{\textwidth}
\scriptsize
\textit{Note:} Mean$\pm$sample SD, $n=3$ coupled source/target seeds. Development F1 is the pinned-floor selection proxy; test-subset F1 is exactly rematched at each frozen threshold and is a box-detection, not patient-diagnosis, endpoint.
\end{minipage}
\vspace{-1ex}
\end{table*}

\begin{figure*}[t]
    \centering
    \includegraphics[width=0.99\linewidth]{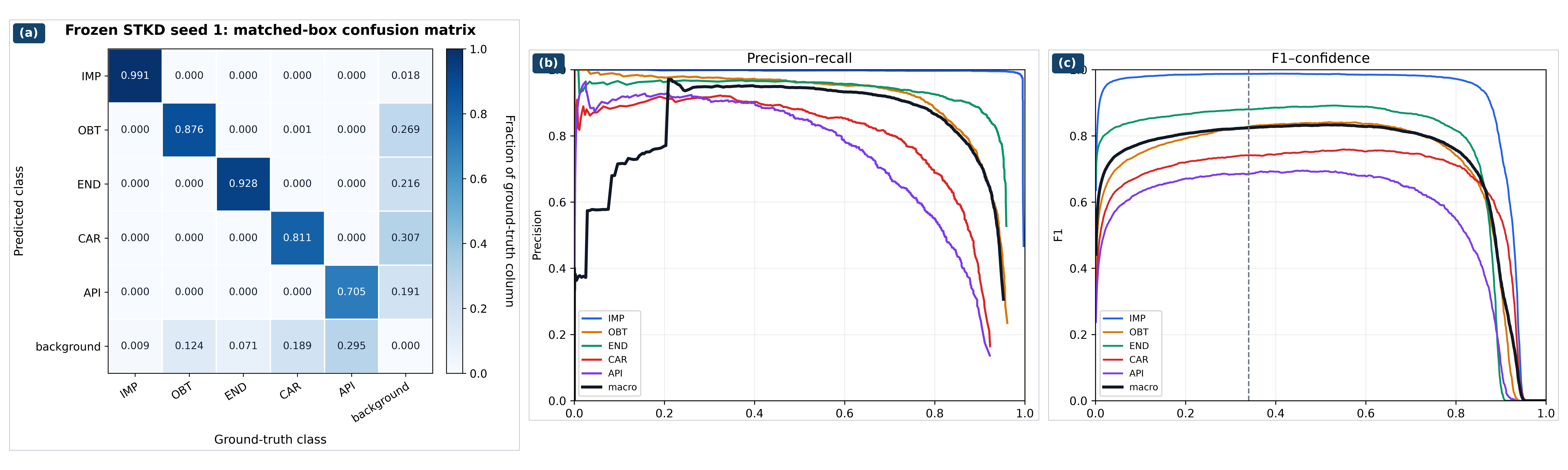}
    \caption{Frozen STKD seed-1 diagnostics on the test subset (confidence 0.340). (a) Normalized confusion matrix at IoU 0.50 (background row: missed boxes; column: unmatched detections). (b) Precision--recall and (c) F1--confidence curves; dashed line: frozen operating point.}
    \label{fig:stkd_diagnostics}
    \vspace{-1ex}
\end{figure*}

\begin{table}[t]
\caption{STKD Seed-1 Class Results on Test Subset}
\label{tab:stkd_classwise_results}
\centering
\begin{minipage}{\columnwidth}
\resizebox{\linewidth}{!}{%
\begin{tabular}{lrrrrrrr}
\toprule
\textbf{Class} & \textbf{TP} & \textbf{FP} & \textbf{FN} & \textbf{Precision} & \textbf{Recall} & \textbf{F1} & \textbf{AP$_\text{50:95}$} \\
\midrule
IMP & 2,304 &  36 & 22  & 0.9846 & 0.9905 & 0.9876 & 0.8094 \\
OBT & 1,938 & 530 & 274 & 0.7853 & 0.8761 & 0.8282 & 0.6831 \\
END & 2,171 & 425 & 168 & 0.8363 & 0.9282 & 0.8798 & 0.6392 \\
CAR & 1,303 & 604 & 304 & 0.6833 & 0.8108 & 0.7416 & 0.5249 \\
API &   753 & 375 & 315 & 0.6676 & 0.7051 & 0.6858 & 0.4471 \\
\bottomrule
\end{tabular}%
}
\vskip 4pt
\scriptsize
Frozen confidence 0.340 and IoU 0.50. TP/FP/FN reproduce the immutable per-image ledger exactly; AP$_\text{50:95}$ is taken from the common evaluator. All endpoints are box-level.
\end{minipage}
\vspace{-3ex}
\end{table}

\section{Results}
\subsection{Source architecture and distillation}
Table~\ref{tab:kd_lineage_results} shows that the source screen separates internal from external ranking. YOLO26-L has the highest internal mAP$_\text{50:95}$ (0.3236), but YOLO26-X is strongest externally (0.2505). RF-DETR-L attains 0.3144 internally and 0.2206 externally; ECDet-X is lower on both. Every family declines externally relative to its internal result. Given unequal budgets and routes, this supports only a family-level observation that internal ranking is unstable under acquisition shift.

Across three validation seeds, B1, STKD, and MTKD reach 0.2236$\pm$0.0065, 0.2244$\pm$0.0033, and 0.2246$\pm$0.0040 mAP$_\text{50:95}$. MTKD's 0.0010 advantage over B1 is smaller than the between-seed SD and is estimated on the duplicated validation split. The selected internal checkpoints favor STKD and MTKD (0.3230, 0.3201 vs 0.3037), but the external checkpoints reverse the ordering. As the MTKD checkpoint (seed 1) is not seed-matched to B1 and STKD (seed 0), these test results cannot reliably rank the three training methods.

The bounded DINOv3/ConvNeXt candidate scores 0.2261 validation mAP$_\text{50:95}$ versus 0.2238 for the matched B1 reference (+0.0023), while mAP$_\text{50}$ falls by 0.0028. It misses the +0.005 advancement gate, so no final source test is run. This changes several things at once, and using the duplicated split does not isolate a DINOv3 effect (Details are in Supplementary S2.4).

\textbf{Source external errors}. AP$_\text{50:95}$ spans 0.003 for SRD to 0.453 for ORD; sparse support, including ORD (n=6), precludes strong rare-class conclusions. Per-class values appear in Supplementary Table S3 and Fig. S1.

\subsection{Cross-dataset specialization}
Before target fine-tuning, five-label zero-shot transfer is weak: the best retained mAP$_\text{50:95}$ is 0.0965 (Supplementary Table S5). Teacher-free specialization raises three-seed development mAP$_\text{50:95}$ to 0.6456$\pm$0.0044 for B1, 0.6408$\pm$0.0123 for STKD, and 0.6348$\pm$0.0060 for MTKD. STKD has a slightly higher development F1 estimate, but B1 leads under the original mAP-based selection rule. This difference led to the documented change in model selection, not a retrospective claim that STKD performs better overall.

In Table~\ref{tab:oralxrays9_means}, B1 has the strongest test subset arm-mean macro F1 (0.8239$\pm$0.0027), micro F1 (0.8449$\pm$0.0035), mAP$_\text{50}$ (0.8541$\pm$0.0014), and mAP$_\text{50:95}$ (0.6130$\pm$0.0067). STKD and MTKD exceed B1 only at mAP$_\text{75}$ (0.6613 and 0.6589 versus 0.6587). MTKD is lowest on macro F1, micro F1, mAP$_\text{50}$ and mAP$_\text{50:95}$. Thus, source KD shows no consistent overall benefit on the target dataset.

Resampling test images gives an STKD--B1 macro-F1 difference of -0.0033 (95\% interval [-0.0064, -0.00008]). As the models were not retrained, this does not establish a reliable difference between training methods.

\begin{table*}[t]
\caption{Development-Set Inference Agreement}
\label{tab:route_fidelity}
\centering
\resizebox{\textwidth}{!}{
\begin{tabular}{lrrrrcc}
\toprule
\textbf{Direct Comparison} & \textbf{Macro-$\text{F}_1$ Drift} & \textbf{Micro-$\text{F}_1$ Drift} & \textbf{Max Class-$\text{F}_1$ Drift} & \textbf{Exact Triplets (\%)} & \textbf{Status} \\
\midrule
Desktop validator $\rightarrow$ Predictor   & 0.00088  & 0.00036 & 0.00469 & 97.59 & \textbf{RED}   \\
Desktop predictor $\rightarrow$ Orin PyTorch & 0.00163  & 0.00098 & 0.01047 & 98.69 & \textbf{RED}   \\
Orin PyTorch $\rightarrow$ TensorRT FP16    & 0.00006 & 0.00030 & 0.00118 & 99.68 & \textbf{GREEN} \\
\bottomrule
\end{tabular}
}
\vspace{1ex}

\begin{minipage}{\textwidth}
\begin{flushleft}
\scriptsize
\textit{Note:} Evaluation performed on development split only. Green gate requires macro/micro-$\text{F}_1$ drift $\le 0.005$, maximum class-$\text{F}_1$ drift $\le 0.010$, exact-triplet rate $\ge 0.995$, and prediction-agreement $\text{F}_1$ at IoU 0.90 $\ge 0.995$. Amber relaxes these limits to 0.010/0.010/0.020/0.990/0.990, respectively; any violation yields Red (worst-endpoint rule). Within-Orin agreement $\text{F}_1$ is 0.99749. Gates assess defined, not clinical-equivalence, thresholds.
\end{flushleft}
\end{minipage}
\vspace{-1ex}
\end{table*}

\subsection{Frozen deployment checkpoint}
After development testing but before viewing final test results, STKD seed 1 and confidence 0.340 were fixed. Repeating the evaluation after viewing the results reproduced the same TP/FP/FN counts for every image in the test subset at IoU 0.50. It obtains macro precision 0.7914, macro recall 0.8622, macro F1 0.8246, micro F1 0.8473, mAP$_\text{50}$ 0.8556, mAP$_\text{75}$ 0.6632, and mAP$_\text{50:95}$ 0.6208. Its fixed-checkpoint 10,000-resample macro-F1 interval is [0.8169,0.8321]. Reinstating the two preidentified duplicates only as a no-tuning descriptive analysis gives macro F1 0.8246 and mAP$_\text{50:95}$ 0.6201, leaving the interpretation unchanged. Full results are in Supplementary S4.2 (Table S10)

Figure~\ref{fig:stkd_diagnostics} and Table~\ref{tab:stkd_classwise_results} show heterogeneous classwise behavior. IMP is easiest (F1 0.9876; AP$_\text{50:95}$ 0.8094). Thresholded F1 ranks END (0.8798) above OBT (0.8282), although AP$_\text{50:95}$ ranks OBT (0.6831) above END (0.6392). CAR reaches F1 0.7416 and AP$_\text{50:95}$ 0.5249. API remains the principal bottleneck (precision 0.6676, recall 0.7051, F1 0.6858, AP$_\text{50:95}$ 0.4471). These class gaps are operationally more informative than the small lineage averages.

Macro and micro aggregation weight the error structure differently. The frozen checkpoint’s micro F1 (0.8473, pooled from the Table~\ref{tab:stkd_classwise_results} counts) exceeds its macro F1 (0.8246), consistent with weaker API and CAR performance under equal class weighting. The exact matched-box matrix contains only three cross-class assignments, versus 1,080 ground-truth boxes mapped to background and 1,967 unmatched detections in the ground-truth-background column. Under this matching policy, missed and spurious boxes, not systematic five-class substitution, dominate the residual error.

% \vspace{-1ex}
\subsection{Edge handoff}
% \vspace{-1ex}
Table~\ref{tab:route_fidelity} checks agreement across three steps of the deployment process using development data and fixed acceptance limits. Within the Orin route, TensorRT FP16 closely reproduces PyTorch: macro-F1 drift is 0.00006, maximum class-F1 drift is 0.00118, 99.68\% of per-image, per-class TP/FP/FN counts are identical, and prediction-agreement F1 at IoU 0.90 is 0.99749. The Orin PyTorch-to-TensorRT comparison passes all checks, but the two earlier comparisons fail. Therefore, agreement across the full desktop-to-device workflow is not established. 

On the recorded Orin Nano Super configuration (JetPack 7.2.1-b49; TensorRT 10.16.2.10), a separate descriptive benchmark repeated a fixed 500-image development sequence three times at static batch-1 and $1280\times1280$. Across 1,500 prediction calls, pooled wall-clock latency was 77.03~ms median and 80.86~ms P95; complete-phase throughput averaged 12.90$\pm$0.06 images/s. Mean total-module VDD$_\text{IN}$ was 17.89$\pm$0.09 W, corresponding to 1.387$\pm$0.013 J/image gross; a 7.34-W engine-resident idle reference gave a secondary signed estimate of 0.819$\pm$0.011 J/image. Peak GPU/T$_j$ temperature was 66.84$^\circ\mathrm{C}$. Measured-phase RAM use peaked at 4.28~GB and process HWM at 1.69~GB; these are unified-memory observations, not an engine footprint.

\section{Discussion}
\subsection{Transfer and evaluation lessons}
Target training reduces the differences between the three source methods. All achieve similar target accuracy, but B1 has the highest macro F1 and mAP$_\text{50:95}$. STKD performs slightly better on the development F1 estimate and test-subset mAP$_\text{75}$. We therefore report both overall accuracy and performance at the selected confidence threshold, without claiming that KD improves overall target accuracy.

Checking image content detects duplicates that filenames miss. Careful weight loading prevents COCO car weights from being assigned to dental CAR, while a common evaluator makes comparisons more consistent. We record class mappings, detection limits, confidence thresholds, model-selection decisions, and device checks so the results can be repeated.

\begin{figure*}
    \centering
    \includegraphics[width=0.99\linewidth]{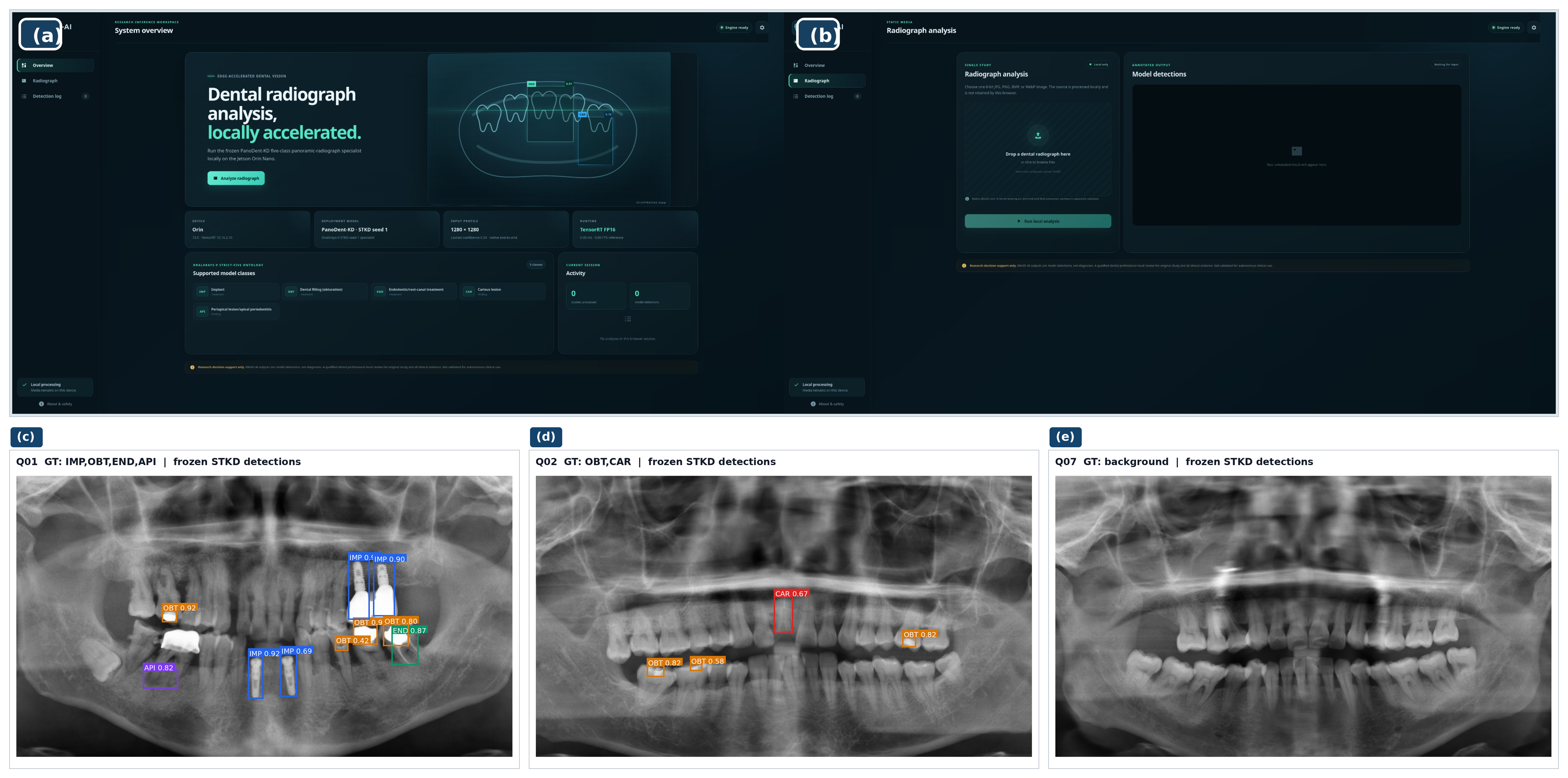}
    \caption{Application workflow and frozen STKD seed-1 examples at confidence 0.340. (a) Dashboard; (b) clinician-controlled review; (c,d) hash-selected, outcome-blind examples (IMP, OBT, END, CAR, API); (e) a background case. Radiographs were manually checked, metadata-stripped, and approved for publication. Research decision support only, not a validated diagnostic device.}
    \label{fig:fig4}
    \vspace{-1ex}
\end{figure*}

\subsection{Application and intended use}
PanoDent-KD is clinician-facing finding localization, not autonomous diagnosis, treatment recommendation, or prognosis. It returns class-coded boxes for IMP, OBT, END, CAR, and API while retaining the human decision point. Its application or intended value is rapid visual indexing of candidate findings; prospective reader studies must determine effects on review efficiency and safety.

\subsection{Application visualization}
Figure~\ref{fig:fig4} joins the intended review interface to approved frozen-checkpoint examples.  Together they show multiclass localization, single-class findings, and an outcome-blind background case rather than a best-case gallery.

At runtime, the fused one-to-one branch returns at most 300 class-scored boxes at the frozen confidence of 0.340. The application maps them to IMP, OBT, END, CAR, and API and keeps acceptance, rejection, and reporting with the clinician. Teachers, KD adapters, and post-test model selection are absent from this deployement.

\section{Limitations}
Limitations include (i) source train--validation duplication; (ii) three coupled seeds; (iii) one five-class target dataset; (iv) unverified patient independence; and (v) one short device session. We did not compare against target-only training or conduct clinical testing. Concurrent use, prolonged heating, and full-workflow latency remain untested. Moreover, the deployment checkpoint was selected through a change not registered in advance.

\section{Conclusion}

This work presents PanoDent-KD, a generalist-to-specialist pipeline for panoramic dental radiograph analysis. On the 2,686-image test subset, all three transfer arms reach approximately 0.82 macro box F1, where B1 leads on arm-level means, while the single frozen STKD seed-1 checkpoint (not an arm mean) reaches 0.8246 and is retained as the deployment checkpoint. This checkpoint's deployment behavior is characterized in detail: exact confusion structure, deterministic evaluation panels, high within-Orin fidelity between PyTorch and TensorRT, 77.03-ms median latency, 12.90 images/s batch-1 throughput, and 1.39 J/image gross module energy. Together, these results report, for one frozen checkpoint, test-subset accuracy and, separately, development-set TensorRT agreement and Orin latency, throughput and energy.

Our contribution is a repeatable workflow connecting source training, exact class transfer, target evaluation, and FP16 deployment. KD teachers are used only during source training and add no inference cost. The dashboard supports clinician review, but the system requires clinical testing before practical use.

For practical use, this evidence is more informative than a single aggregate score: (i) IMP reaches the highest F1 (0.9876); (ii) OBT and END reach F1 0.8282 and 0.8798; and (iii) CAR and API findings define concrete curation priorities for future data collection. As the model, decision threshold, class map, and target-built engine are all fixed, prospective workflow testing can evaluate this exact inspected system without retraining or post-hoc threshold reselection.

{\small
\bibliographystyle{ieeenat_fullname}
\bibliography{main}
}

\clearpage
\renewcommand{\thetable}{S\arabic{table}}
\renewcommand{\thefigure}{S\arabic{figure}}
\setcounter{table}{0}
\setcounter{figure}{0}

\twocolumn[
\centering
\large \textbf{PanoDent-KD: From Generalist Knowledge Distillation to Cross-Dataset Specialist Detection and Embedded Deployment in Panoramic Radiographs}\\
\vspace{0.1cm}
\small \textbf{Supplementary Material}
\vspace{1cm}
]

The supplementary reports aggregate, classwise, and checksum-bound evidence for the source architecture screens, cross-dataset transfers, specialist lineages, and system performance benchmarks. All detection endpoints are box-level rather than patient-level. Mean$\pm$SD denotes the arithmetic mean and sample standard deviation across three coupled seeds.

% 2.1
\begin{table*}[!t]
\centering
\caption{Source architecture screen and selected KD checkpoint reports.}
\small
\resizebox{\textwidth}{!}{%
\begin{tabular}{ll rr l}
\toprule
\textbf{Study} & \textbf{Model} & \textbf{Internal mAP$_\text{50-95}$} & \textbf{External mAP$_\text{50-95}$} & \textbf{Qualification} \\
\midrule
Screen & ECDet-X & 0.2122 & 0.1488 & 50 epochs \\
Screen & RF-DETR-L & 0.3144 & 0.2206 & early stop, epoch 29/100 \\
Screen & RT-DETR-L & 0.3054 & 0.2107 & class remapping disabled \\
Screen & YOLO26-M & 0.3132 & 0.2286 & 75 epochs \\
Screen & YOLO26-L & 0.3236 & 0.2245 & 75 epochs \\
Screen & YOLO26-X & 0.3178 & 0.2505 & 75 epochs \\
Screen & YOLOv8x & 0.2981 & 0.2285 & 1024 pixels, 25 epochs \\
\hline
Selected source report & Seed 0: B1 & 0.3037 & 0.2284 & partly unmatched selected checkpoint \\
Selected source report & Seed 0: STKD & 0.3230 & 0.2215 & partly unmatched selected checkpoint \\
Selected source report & Seed 1: MTKD & 0.3201 & 0.2164 & partly unmatched selected checkpoint \\
\bottomrule
\end{tabular}%
}
\end{table*}

\section*{S1. Training and Evaluation sets}

The source task is 12-class panoramic detection on an ADCD-derived split; the target task is five-class detection on OralXrays-9. B1 denotes supervised source training, STKD source training with one frozen teacher, and MTKD source training with two. All three source arms use the same YOLO26m student. On the target dataset, each student is fine-tuned without a teacher. B1 is therefore a supervised source-to-target baseline, not a target-only baseline.

\textbf{Development set}. The 2,000 target development images were used for model and threshold selection, the zero-shot comparisons, and the device-agreement checks. The original source split contains 42 identical-image groups spanning training and validation. Exact-content checks found no such overlap between target training and development. Patient independence of this derived split could not be established.

\textbf{Primary test subset and full-set sensitivity}. Before viewing the final results, blinded review of potential duplicates identified the presence of two identical test radiographs. The primary test subset, therefore, then contains the remaining 2,686 images and 9,552 annotations for the five classes. A subsequent sensitivity analysis uses the original, i.e., 2,688 images and 9,558 boxes, under the same model and settings. These two exclusions concern the target test, not the 42 source duplicate groups.

\section*{S2. Source Models and Ablations}

\subsection*{S2.1. Architecture Screen}
The detector-family screen used one compatible evaluator, but its training budgets, resolutions, and runtime routes were unequal. It is therefore an empirical architecture screen rather than a controlled compute or latency ablation. The retained RT-DETR-L result disables class remapping; an older name-matched row is excluded because it mapped the pretrained COCO car classifier into dental CAR.

The internal and external rankings differ: YOLO26-L leads the internal screen, whereas YOLO26-X leads externally. Every screened family declines externally. Runtime fields retained in the evidence are route-specific observations and are not compared here.

% 2,2
\begin{table}[H]
\centering
\caption{Source-validation three-seed results.}
\small
\resizebox{\columnwidth}{!}{%
\begin{tabular}{lrrrr}
\toprule
\textbf{Arm} & \textbf{Seed 0} & \textbf{Seed 1} & \textbf{Seed 2} & \textbf{Mean$\pm$SD} \\
\midrule
B1 & 0.2237 & 0.2301 & 0.2171 & 0.2236{\scriptsize$\pm$0.0065} \\
STKD & 0.2265 & 0.2206 & 0.2262 & 0.2244{\scriptsize$\pm$0.0033} \\
MTKD & 0.2268 & 0.2271 & 0.2200 & 0.2246{\scriptsize$\pm$0.0040} \\
\bottomrule
\end{tabular}%
}
\end{table}

\subsection*{S2.2. Three-seed B1/STKD/MTKD Source Ablation}
B1, STKD, and MTKD use the same YOLO26m student and source training recipe across seeds 0-2. Source validation contains 42 exact-image groups shared with source training; these results document the executed experiment but do not support a new state-of-the-art claim.

MTKD exceeds B1 by 0.0010 in mean validation mAP$_\text{50-95}$, less than the observed between-seed dispersion. Selected external reports reverse this ordering (B1 0.2284, STKD 0.2215, MTKD 0.2164), but the selected rows are partly unmatched (MTKD is seed 1; B1 and STKD are seed 0), so these test results cannot reliably rank the three training methods.

\begin{table}[H]
\centering
\caption{Selected MTKD external source checkpoint: classwise evaluator output.}
\label{Suppl_Table_S3}
\small
\resizebox{\columnwidth}{!}{%
\begin{tabular}{lrrrr}
\toprule
\textbf{Class} & \textbf{Instances} & \textbf{F1 derived} & \textbf{AP$_\text{50}$} & \textbf{AP$_\text{50-95}$} \\
\midrule
IMP &  16 & 0.6475 & 0.7176 & 0.3471 \\
PRR & 206 & 0.6063 & 0.5404 & 0.3168 \\
OBT & 897 & 0.6428 & 0.6135 & 0.2589 \\
END & 527 & 0.7081 & 0.6900 & 0.3163 \\
CAR & 296 & 0.3585 & 0.2986 & 0.1168 \\
BON & 680 & 0.1404 & 0.1261 & 0.0348 \\
IMT & 111 & 0.7527 & 0.7284 & 0.3985 \\
API & 106 & 0.3387 & 0.2750 & 0.0895 \\
ROT &  93 & 0.5617 & 0.5460 & 0.2042 \\
FUR &  19 & 0.3021 & 0.2013 & 0.0574 \\
ORD &   6 & 0.5409 & 0.7502 & 0.4529 \\
SRD &   5 & 0.0000 & 0.0041 & 0.0033 \\
\bottomrule
\end{tabular}%
}
\end{table}

% 2.3
\begin{figure*}[!t]
  \centering
  \includegraphics[trim={0 0 0 4cm}, clip, width=0.9\textwidth]{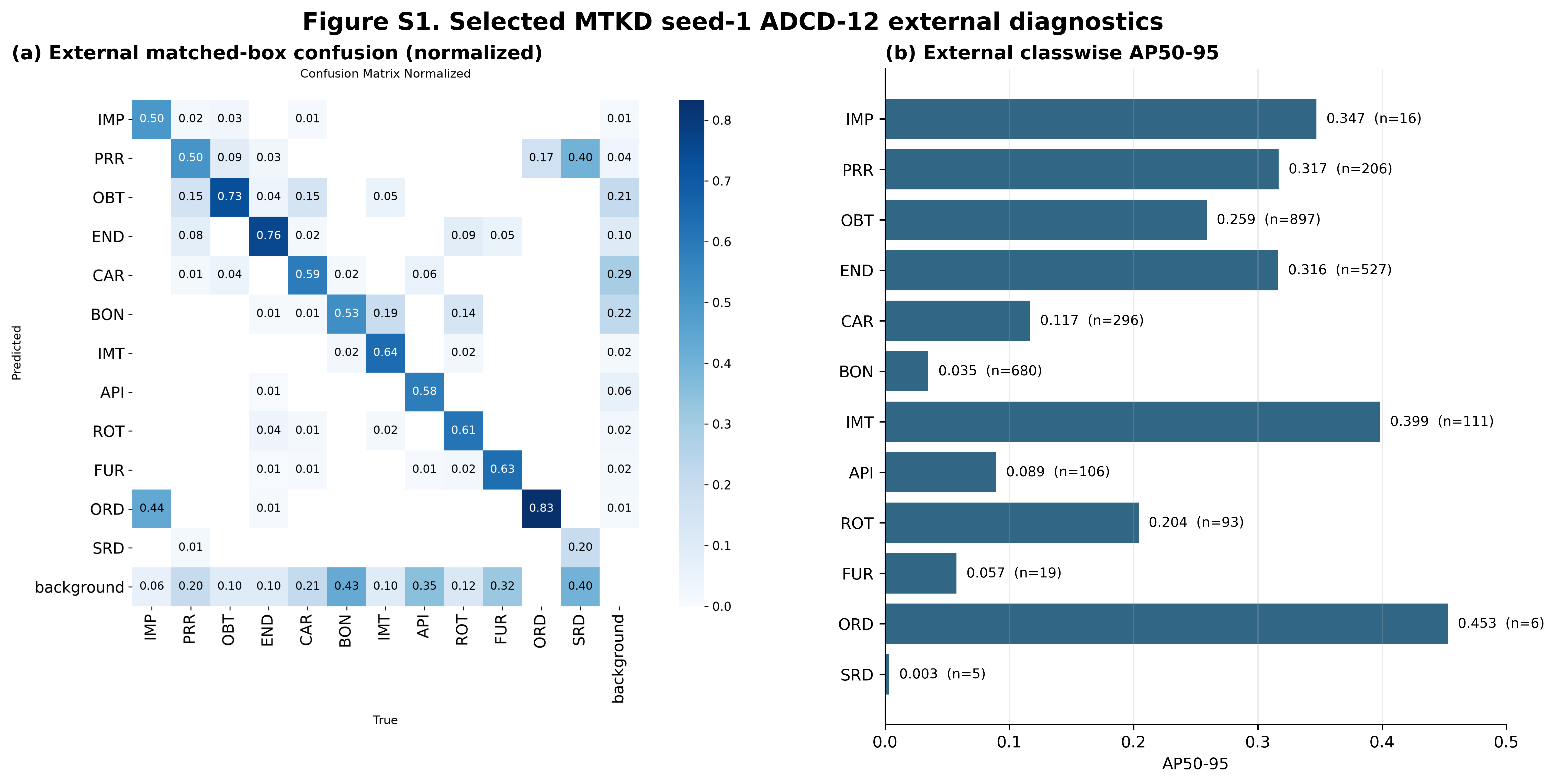}
  \vspace{-2ex}
  \caption{External errors for the selected ADCD-12 MTKD seed-1 checkpoint.}
  \label{Suppl_Fig_S1}
\end{figure*}

\subsection*{S2.3. Class-wise External Results}

The retained class-wise evaluator CSV gives mAP$_\text{50-95}$ values of 0.3471 for IMP and 0.0033 for SRD. 
% These replace the prose values 0.422 and 0.004 in the current main-paper callout. This is a provenance correction, 
This is not a rerun or a model change. ORD has mAP$_\text{50-95}$ 0.4529 but only six instances; it is displayed for completeness and is not treated as a stable upper-bound estimate. Table~\ref{Suppl_Table_S3} reports all classes, and Fig.~\ref{Suppl_Fig_S1} presents the corresponding errors.

% 2.4
\begin{table}[H]
\centering
\caption{Bounded DINOv3-guided candidate versus the matched B1 reference.}
\label{Suppl_Table_S4}
\small
\resizebox{\columnwidth}{!}{%
\begin{tabular}{lrrrrr}
\toprule
\textbf{Row} & \textbf{Precision} & \textbf{Recall} & \textbf{F1} & \textbf{mAP$_\text{50}$} & \textbf{mAP$_\text{50-95}$} \\
\midrule
Matched B1 ref. & 0.5308 & 0.5146 & 0.5226 & 0.5013 & 0.2238 \\
DINOv3 cand. & 0.5472 & 0.5027 & 0.5240 & 0.4985 & 0.2261 \\
Candidate$-$B1 & +0.0164 & $-$0.0118 & +0.0015 & $-$0.0028 & +0.0023 \\
\bottomrule
\end{tabular}%
}
\end{table}

% 2.4
\begin{figure*}[!t]
  \centering
  \includegraphics[trim={0 0 0 3.5cm}, clip, width=0.9\textwidth]{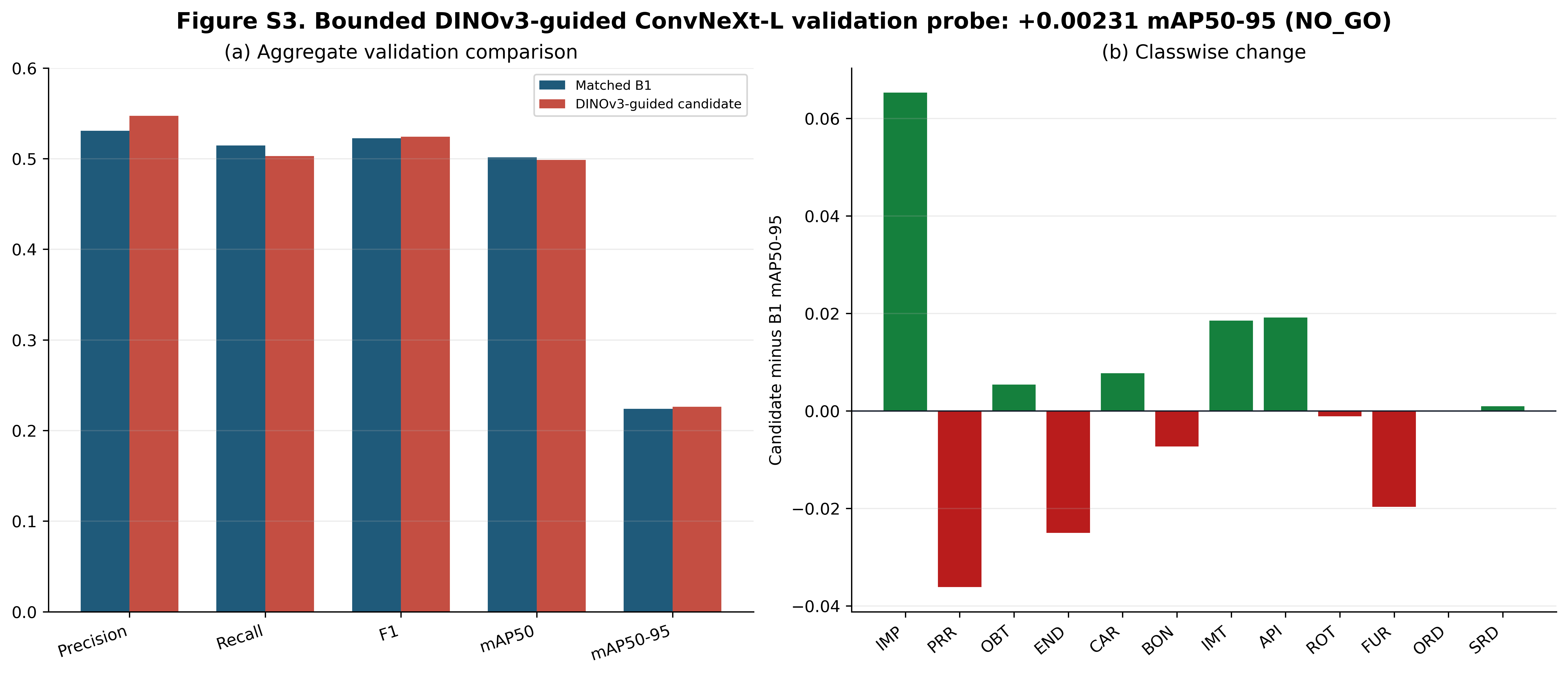}
  \vspace{-2ex}
  \caption{Bounded DINOv3-guided ConvNeXt-L validation probe.}
  \label{Suppl_Fig_S2}
\end{figure*}

\subsection*{S2.4. Bounded DINOv3/ConvNeXt-L Probe}
A DINOv3/ConvNeXt-L candidate was compared with its matched supervised reference in a seed-0, 25-epoch validation probe, as in Table~\ref{Suppl_Table_S4} and Fig.~\ref{Suppl_Fig_S2}. Its increase in mAP50:95 (+0.002309) fell below the prespecified +0.005 criterion, while mAP50 decreased. No source test was run for this candidate. The probe changed several components together, so it cannot isolate a DINOv3 effect.

\textbf{Class abbreviations}. The specialist uses IMP (implant), OBT (dental filling), END (root canal filling), CAR (caries), and API (apical periodontitis). The original ADCD-12 source codes are retained in Table~\ref{Suppl_Table_S3} to match the evaluator. Sparse-class estimates should be read with the corresponding box counts rather than as evidence of stable clinical performance.

% 3.1
\begin{table*}[!t]
\centering
\caption{Zero-shot shared-five-class development results.}
\label{Suppl_Table_S5}
\small
\resizebox{\textwidth}{!}{%
\begin{tabular}{llrrrrrrr}
\toprule
\textbf{Rank} & \textbf{Model} & \textbf{Seed} & \textbf{Precision} & \textbf{Recall} & \textbf{Derived F1} & \textbf{mAP$_\text{50}$} & \textbf{mAP$_\text{75}$} & \textbf{mAP$_\text{50-95}$} \\
\midrule
1 & YOLO26M-STKD & 0 & 0.3723 & 0.4082 & 0.3894 & 0.3022 & 0.0376 & 0.0965 \\
2 & YOLO26X & 0 & 0.3744 & 0.4148 & 0.3936 & 0.2982 & 0.0454 & 0.0948 \\
3 & YOLO26M-B1 & 1 & 0.4041 & 0.4284 & 0.4159 & 0.3179 & 0.0385 & 0.0908 \\
4 & YOLOv8X & 0 & 0.4114 & 0.4305 & 0.4207 & 0.3112 & 0.0284 & 0.0828 \\
5 & YOLO26M-MTKD & 1 & 0.3120 & 0.3928 & 0.3478 & 0.2316 & 0.0192 & 0.0624 \\
\bottomrule
\end{tabular}%
}
\end{table*}

\section*{S3. Cross-dataset Transfer: Transfer to Five Target Classes}

\subsection*{S3.1. Before Target Fine-tuning}
Before target fine-tuning in Table~\ref{Suppl_Table_S5}, source checkpoints were evaluated on the 2,000-image target development set after retaining the five shared class rows. These are development-only transfer results.

Zero-shot mAP$_\text{50-95}$ remains below 0.10 for every retained checkpoint, motivating teacher-free target specialization.

% 3.2
\begin{figure*}[!t]
  \centering
  \includegraphics[width=0.9\textwidth]{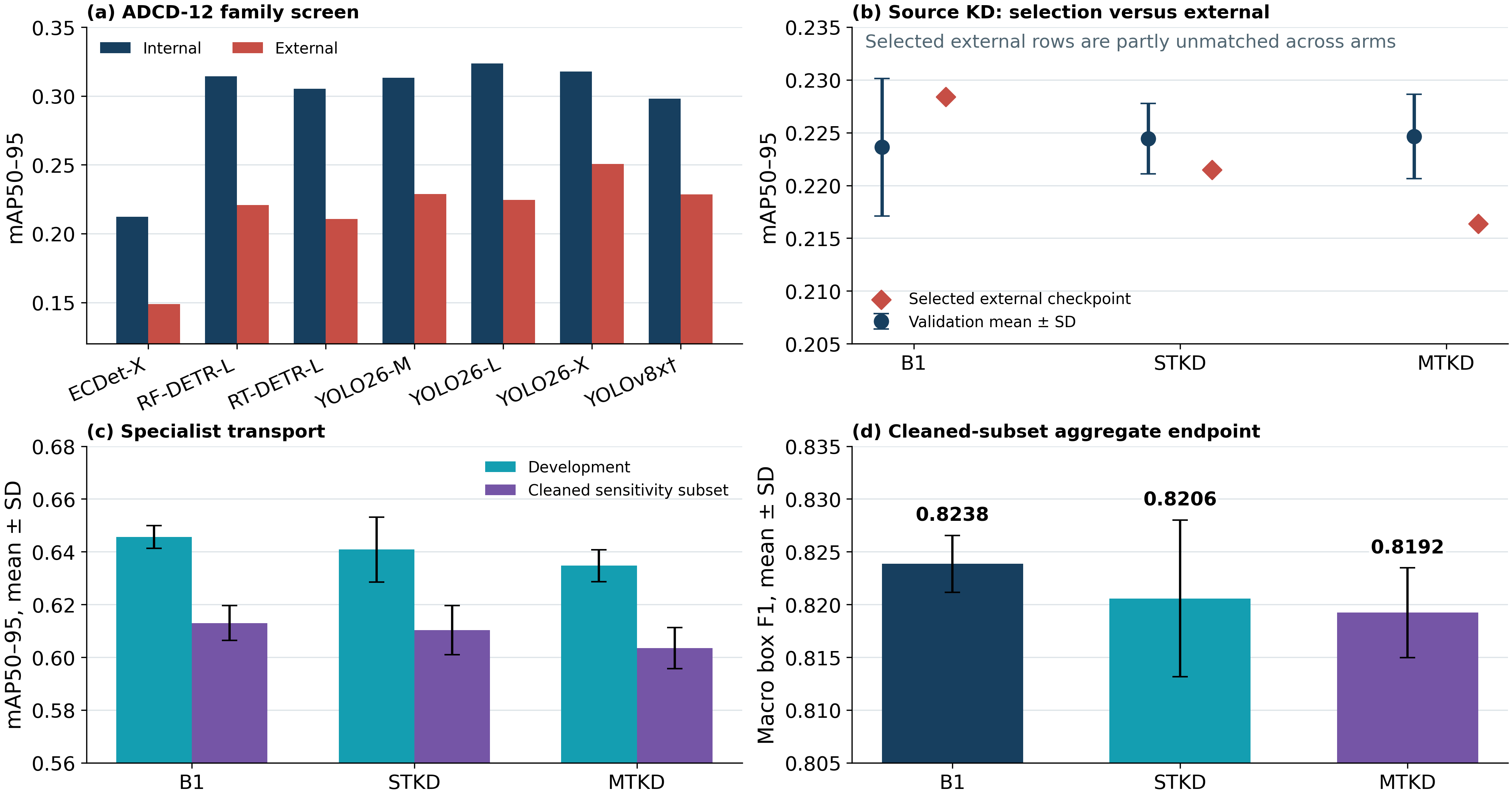}
  \caption{Source model screen, source training arms, target development, and primary test endpoints. Source validation means using three seeds. Target arm means using three teacher-free specialist runs per arm.}
  \label{Suppl_Fig_S3}
\end{figure*}

% 3.2
\begin{table*}[!t]
\centering
\caption{Target results across three seeds. The first two columns use the development set and the remaining use the 2,686-image test-subset: Values are mean$\pm$sample SD.}
\small
\resizebox{\textwidth}{!}{%
\begin{tabular}{lrrrrrrr}
\toprule
\textbf{Arm} & \textbf{Dev. proxy F1} & \textbf{Dev. mAP$_\text{50-95}$} & \textbf{Test macro F1} & \textbf{Test micro F1} & \textbf{Test mAP$_\text{50}$} & \textbf{Test mAP$_\text{75}$} & \textbf{Test mAP$_\text{50-95}$} \\
\midrule
B1 & 0.8410{\scriptsize$\pm$0.0023} & 0.6456{\scriptsize$\pm$0.0044} & 0.8239{\scriptsize$\pm$0.0027} & 0.8449{\scriptsize$\pm$0.0035} & 0.8541{\scriptsize$\pm$0.0014} & 0.6587{\scriptsize$\pm$0.0021} & 0.6130{\scriptsize$\pm$0.0067} \\
STKD & 0.8428{\scriptsize$\pm$0.0064} & 0.6408{\scriptsize$\pm$0.0123} & 0.8206{\scriptsize$\pm$0.0074} & 0.8418{\scriptsize$\pm$0.0086} & 0.8539{\scriptsize$\pm$0.0019} & 0.6613{\scriptsize$\pm$0.0019} & 0.6103{\scriptsize$\pm$0.0094} \\
MTKD & 0.8402{\scriptsize$\pm$0.0021} & 0.6348{\scriptsize$\pm$0.0060} & 0.8192{\scriptsize$\pm$0.0043} & 0.8404{\scriptsize$\pm$0.0041} & 0.8495{\scriptsize$\pm$0.0039} & 0.6589{\scriptsize$\pm$0.0031} & 0.6035{\scriptsize$\pm$0.0078} \\
\bottomrule
\end{tabular}%
}
\end{table*}

% 3.2
\begin{table*}[!t]
\centering
\caption{Patch group and reproduced evidence map.}
\small
\resizebox{\textwidth}{!}{%
\begin{tabular}{ll}
\toprule
\textbf{Patch group} & \textbf{Reproduced evidence} \\
\midrule
Source KD training & B1/STKD/MTKD recipes, feature-tap adapters, seed control, teacher/adaptor exclusion from serialized student \\
Specialist transfer and training & Exact five-row mapping in both classifier branches, transfer tensor audit, nine teacher-free fine-tunes \\
Frozen five-class development evaluation & Unsupported-class masking before top-k selection and shared-class validator allowlisting \\
Aggregate-evidence rendering & Matched confusion matrices, PR/F1 curves, 2,688 no-tuning sensitivity, inference agreement, and Orin summaries \\
Orin descriptive benchmark & Synchronized latency and telemetry aggregation under the recorded static FP16 engine contract \\
\bottomrule
\end{tabular}%
}
\end{table*}

\subsection*{S3.2. Teacher-free Target Specialization}
Five exact source classifier rows initialize five target rows in both classifier branches at all three scales: Classifier rows [0,2,3,4,7] in ADCD-12 map to target rows [0,1,2,3,4]. The audit for each source checkpoint recorded 756 loaded tensors and 12 remapped classifier tensors; none were skipped or newly initialized. Optimizer state was not transferred. All nine initialized students underwent 75 epochs of supervised target training, using the same seed as their source model.

Development proxy F1 is recomputed from the pinned confidence-floor statistics and is not the validator-derived F1 field. B1 leads the prespecified test-subset aggregate endpoints except mAP$_\text{75}$, where STKD (+0.0026) and MTKD (+0.0003) are higher. The target results do not show an aggregate advantage from source KD.

The development-set paired image bootstrap for mAP$_\text{50-95}$ gives STKD$-$B1 = $-$0.0049 with a 95\% interval [$-0.0073$, $-0.0023$] across the fixed checkpoints. On the 2,686-image test-subset, the STKD$-$B1 arm-mean macro-F1 point difference is $-$0.0033; the bootstrap median is $-$0.0033 with a 95\% interval [$-0.0064$, $-0.0001$]. Both analyses condition on fixed checkpoints and images and omit retraining uncertainty; neither is a method-level significance test. Fig.~\ref{Suppl_Fig_S3} summarizes the comparisons.

% 4.0
\begin{table*}[!t]
\centering
\small
\caption{Frozen STKD seed-1 classwise results, 2,686 test-subset images. TP/FP/FN, precision, recall, and F1 use confidence 0.340 and IoU 0.50. AP integrates ranked detections at the indicated IoU values.}
\label{Suppl_Table_S8}
% \begin{adjustbox}
\begin{tabular}{lrrrr|rrr|rrr}
\hline
\textbf{Class} & \textbf{Targets} & \textbf{TP} & \textbf{FP} & \textbf{FN} & \textbf{Precision} & \textbf{Recall} & \textbf{F1} & \textbf{AP$_\text{50}$} & \textbf{AP$_\text{75}$} & \textbf{AP$_\text{50-95}$} \\
\hline
IMP & 2,326 & 2,304 & 36 & 22 & 0.9846 & 0.9905 & 0.9876 & 0.9929 & 0.8799 & 0.8094 \\
OBT & 2,212 & 1,938 & 530 & 274 & 0.7853 & 0.8761 & 0.8282 & 0.8898 & 0.7348 & 0.6831 \\
END & 2,339 & 2,171 & 425 & 168 & 0.8363 & 0.9282 & 0.8798 & 0.9053 & 0.6826 & 0.6392 \\
CAR & 1,607 & 1,303 & 604 & 304 & 0.6833 & 0.8108 & 0.7416 & 0.7677 & 0.5531 & 0.5249 \\
API & 1,068 & 753 & 375 & 315 & 0.6676 & 0.7051 & 0.6858 & 0.7224 & 0.4653 & 0.4471 \\
\hline
\end{tabular}
% \end{adjustbox}
\end{table*}

% 4.0
\begin{figure*}[!t]
  \centering
  \includegraphics[width=0.9\textwidth]{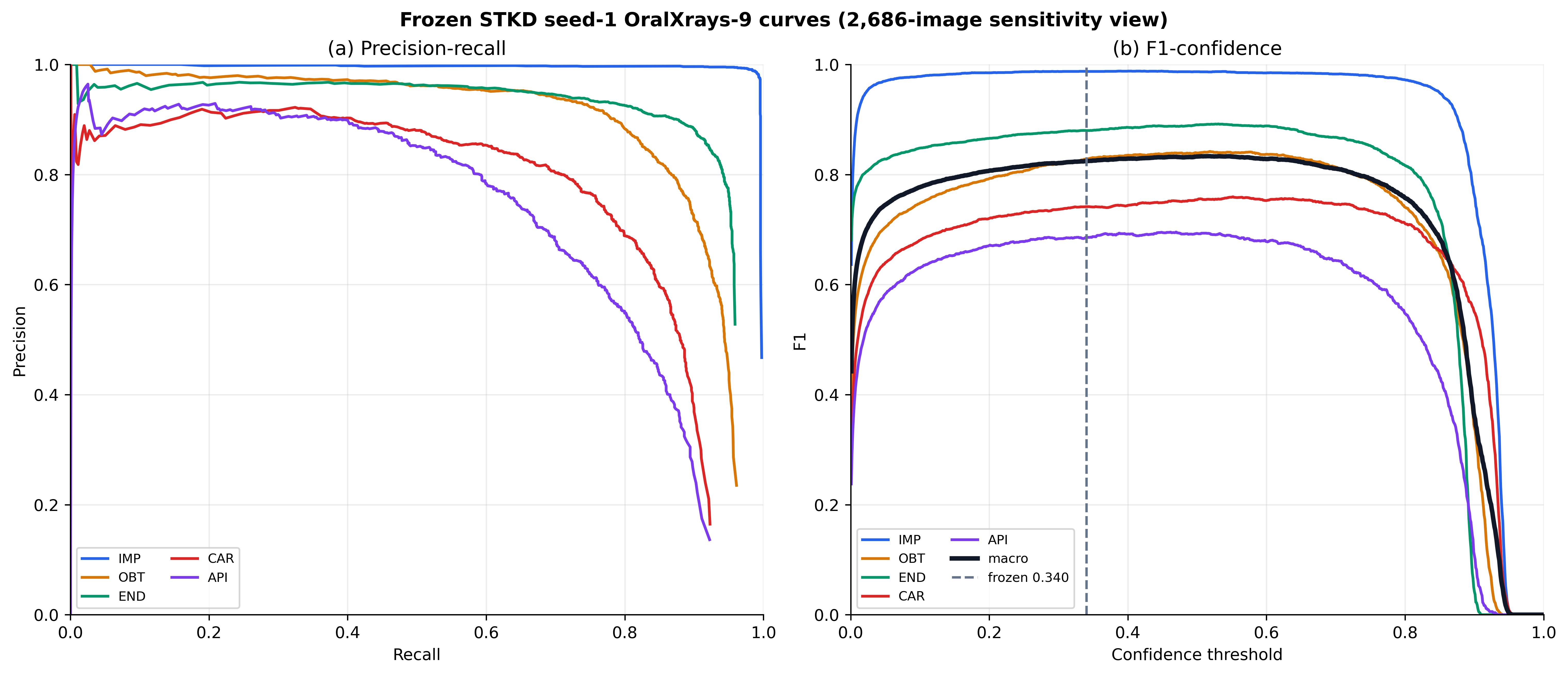}
  \vspace{-2ex}
  \caption{Five-class precision–recall and F1–confidence curves on the primary test subset. The vertical marker denotes the development-frozen confidence of 0.340.}
\end{figure*}

\section*{S4. Frozen Deployment Checkpoint: STKD Seed-1}
The deployment checkpoint and confidence 0.340 were fixed after development and before final-test results were viewed, through a dated change that was not registered in advance. Repeating the evaluation after viewing the results reproduced every per-image TP/FP/FN count on the 2,686-image test subset. Aggregate results are: macro precision 0.7914, macro recall 0.8622, macro F1 0.8246, micro precision 0.8113, micro recall 0.8866, micro F1 0.8473, mAP$_\text{50}$ 0.8556, mAP$_\text{75}$ 0.6632, and mAP$_\text{50-95}$ 0.6208. A 10,000-resample fixed-checkpoint image bootstrap gives a macro-F1 median of 0.8246 and 95\% interval [$0.8169$, $0.8321$].

% 4.1
\begin{figure*}[!t]
  \centering
  \begin{minipage}[b]{0.48\textwidth}
    \centering
    \includegraphics[width=\linewidth]{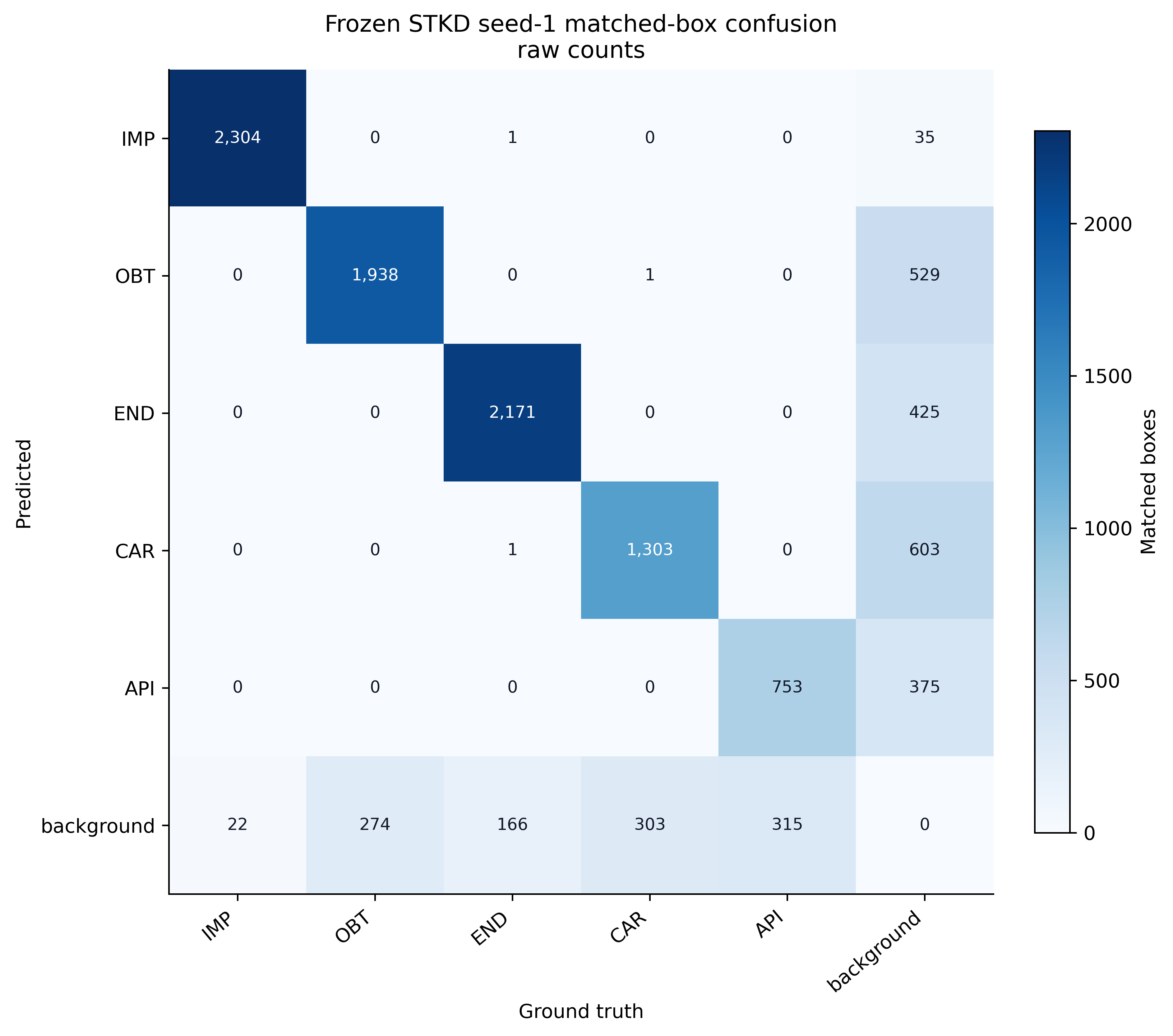}
    \captionof{figure}{Raw matched-box confusion matrix for frozen STKD seed-1 on the 2,686-image test-subset.}
    \label{Suppl_Fig_S5}
  \end{minipage}
  \hfill
  \begin{minipage}[b]{0.48\textwidth}
    \centering
    \includegraphics[width=\linewidth]{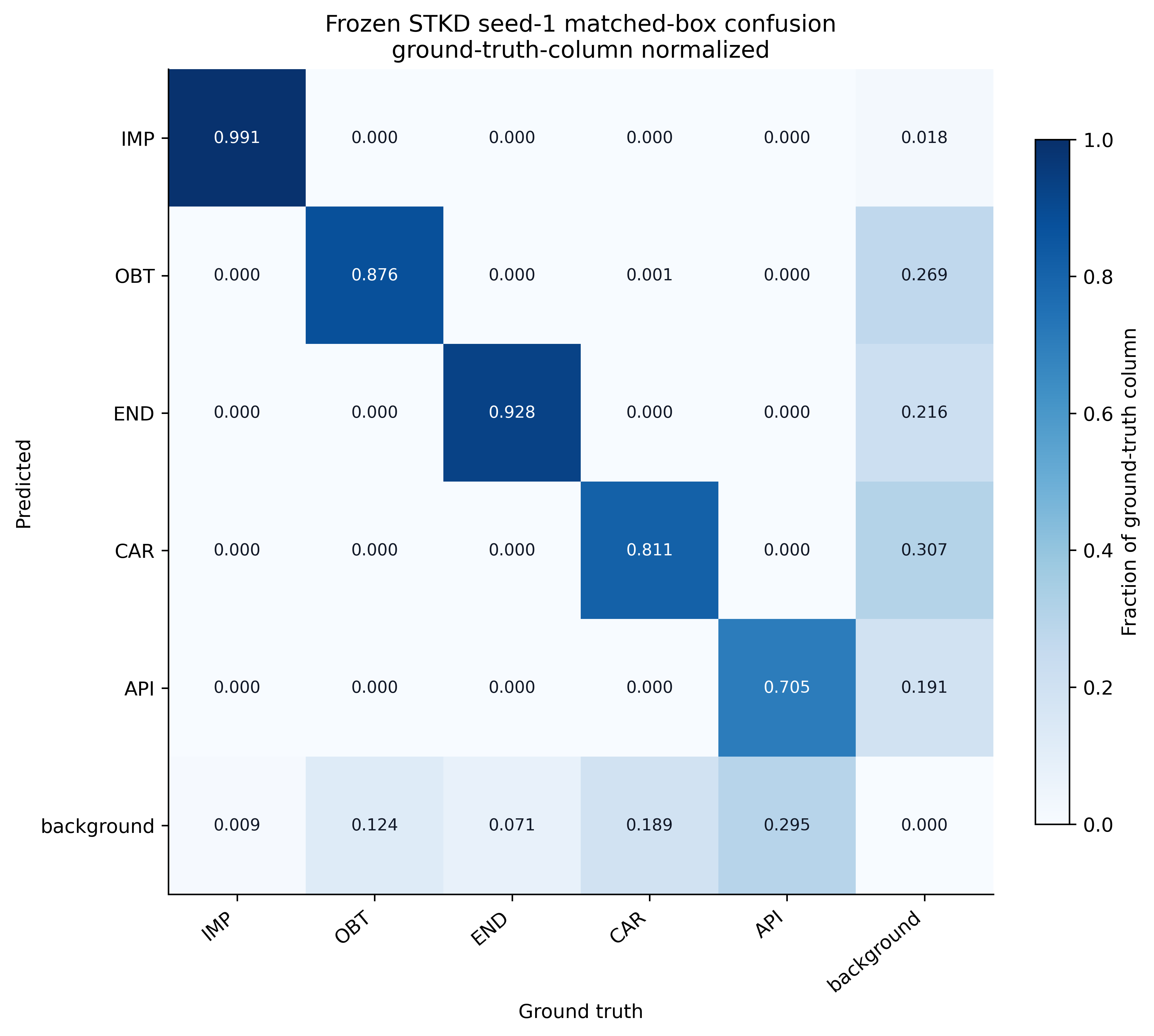}
    \captionof{figure}{Normalized Ground-truth-column of the matched-box confusion matrix in Fig.~\ref{Suppl_Fig_S5}}
    \label{Suppl_Fig_S6_Normalized_S5}
  \end{minipage}
\end{figure*}

% 4.1
\begin{table}
\centering
\caption{Raw matched-box confusion matrix for the frozen STKD seed-1 checkpoint.}
\label{Suppl_Table_S9}
\small
\resizebox{\columnwidth}{!}{%
\begin{tabular}{lrrrrrr}
\toprule
\textbf{Pred.} $\backslash$ \textbf{GT} & \textbf{IMP} & \textbf{OBT} & \textbf{END} & \textbf{CAR} & \textbf{API} & \textbf{Bkg.} \\
\midrule
IMP & 2,304 & 0 & 1 & 0 & 0 & 35 \\
OBT & 0 & 1,938 & 0 & 1 & 0 & 529 \\
END & 0 & 0 & 2,171 & 0 & 0 & 425 \\
CAR & 0 & 0 & 1 & 1,303 & 0 & 603 \\
API & 0 & 0 & 0 & 0 & 753 & 375 \\
Bkg. & 22 & 274 & 166 & 303 & 315 & -- \\
\bottomrule
\end{tabular}%
}
\end{table}

\subsection*{S4.1. Exact Matched-box Confusion Structure}
Rows, as in Table~\ref{Suppl_Table_S9}, are predicted classes and columns are ground-truth classes. The ground-truth-background column contains unmatched detections; the predicted-background row contains missed ground-truth boxes.

Only three matched detections are assigned to another target class: one END $\rightarrow$ IMP, one CAR $\rightarrow$ OBT, and one END $\rightarrow$ CAR under the table orientation above (arrows denote ground truth $\rightarrow$ prediction). The dominant residual errors are 1,080 missed ground-truth boxes and 1,967 unmatched detections, not systematic cross-class substitution.

% 4.2
\begin{table*}[!t]
\centering
\caption{Frozen checkpoint results on the primary subset (2,686 images) versus full set (2,688 images). Differences are full minus primary, calculated before rounding.}
\label{Suppl_Table_S10}
\small
\resizebox{\textwidth}{!}{%
\begin{tabular}{lrrrrrrrrrr}
\toprule
\textbf{Scope} & \textbf{Macro Prec.} & \textbf{Macro Rec.} & \textbf{Macro F1} & \textbf{Micro Prec.} & \textbf{Micro Rec.} & \textbf{Micro F1} & \textbf{mAP$_\text{50}$} & \textbf{mAP$_\text{75}$} & \textbf{mAP$_\text{50-95}$} \\
\midrule
Test set (2,686) & 0.791396 & 0.862146 & 0.824602 & 0.811285 & 0.886621 & 0.847281 & 0.855626 & 0.663159 & 0.620752 \\
Full (2,688) & 0.791417 & 0.862123 & 0.824604 & 0.811297 & 0.886587 & 0.847273 & 0.855468 & 0.662254 & 0.620080 \\
\midrule
Difference & +0.000021 & $-$0.000023 & +0.000002 & +0.000013 & $-$0.000033 & $-$0.000008 & $-$0.000158 & $-$0.000905 & $-$0.000672 \\
\bottomrule
\end{tabular}%
}
\end{table*}

% 4.2
\begin{figure*}[!t]
  \centering
  \includegraphics[trim={0 0 0 3.5cm}, clip, width=0.9\textwidth]{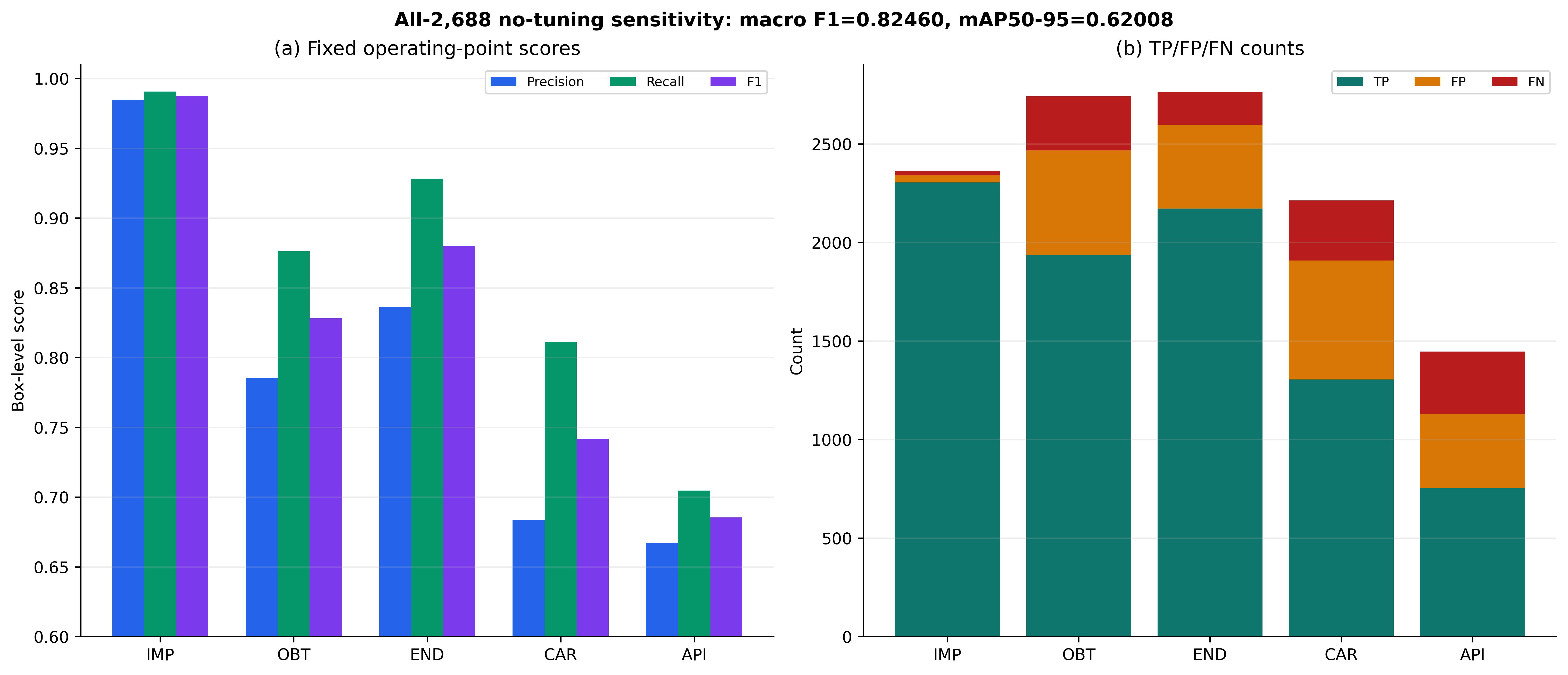}
  \vspace{-2ex}
  \caption{All-2,688 sensitivity analysis: This analysis retains the same frozen checkpoint, class mapping, threshold, and evaluation settings.}
  \label{Suppl_Fig_S7}
\end{figure*}

\subsection*{S4.2. All-2,688 images: Sensitivity Analysis}
The two duplicate-adjudicated cases were added back only after final-test results had been viewed. The checkpoint, class map, confidence 0.340, and IoU 0.50 remained unchanged, i.e., there was no tuning or reselection. Table~\ref{Suppl_Table_S10} and Fig.~\ref{Suppl_Fig_S7} give the sensitivity analysis.

The 2,688-image classwise TP/FP/FN values are IMP 2,305/36/22, OBT 1,938/530/274, END 2,172/425/168, CAR 1,305/604/304, and API 754/376/316. The negligible aggregate changes leave the interpretation unchanged.

% 5.1
\begin{figure*}[!t]
  \centering
  \includegraphics[trim={0 0 0 3.65cm}, clip, width=0.9\textwidth]{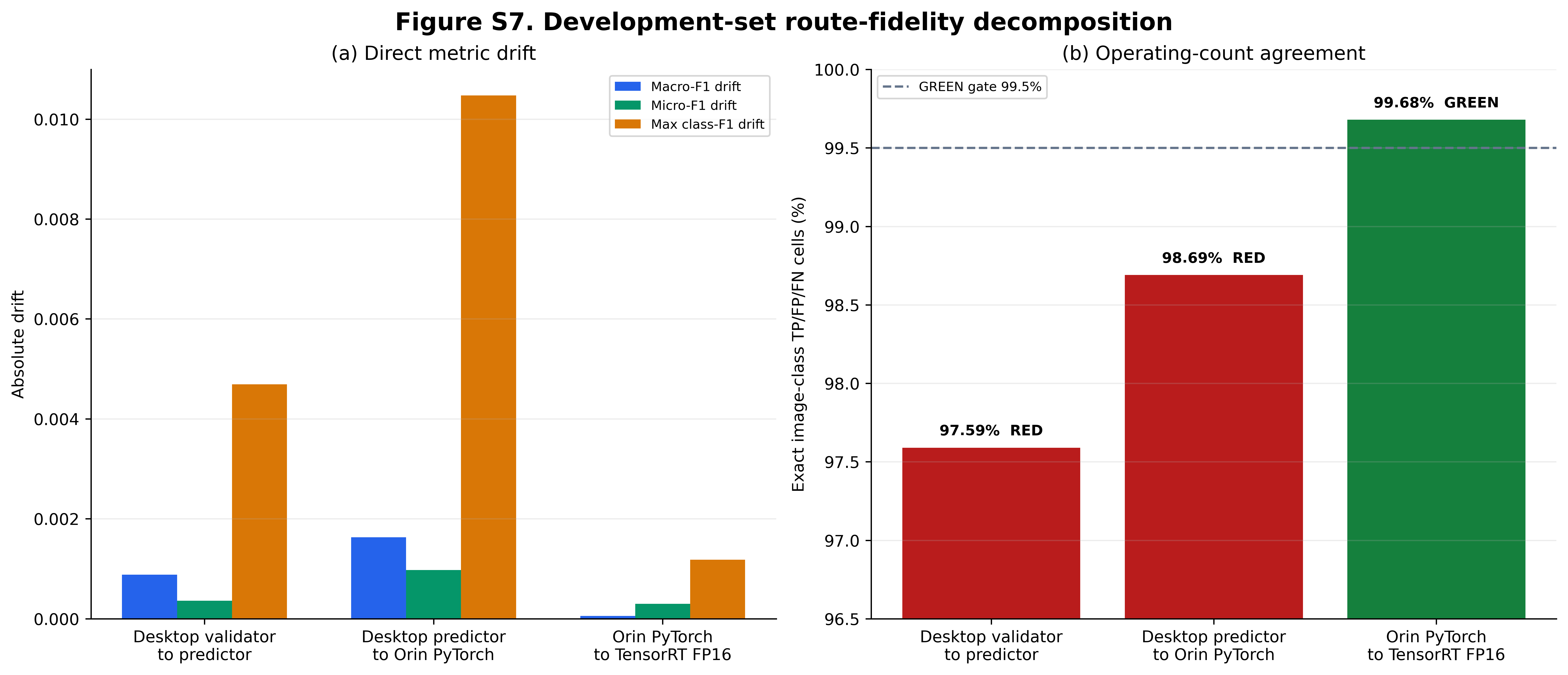}
  \vspace{-2ex}
  \caption{The three development-set comparisons in Table~\ref{Suppl_Table_S11}. Only the within-Orin PyTorch-to-TensorRT step meets all prescribed thresholds.}
\end{figure*}

% 5.1
\begin{table*}[!t]
\centering
\caption{Direct inference-agreement comparisons on the full target development set. Drifts are absolute differences in F1. The count column reports per-image, per-class TP/FP/FN triplets.}
\label{Suppl_Table_S11}
\small
\resizebox{\textwidth}{!}{%
\begin{tabular}{lrrrcl}
\toprule
\textbf{Direct comparison} & \textbf{Macro-F1 drift} & \textbf{Micro-F1 drift} & \textbf{Max class-F1 drift} & \textbf{Exact Count} & \textbf{Status} \\
\midrule
Desktop validator $\rightarrow$ Desktop predictor & 0.000881 & 0.000360 & 0.004690 & 97.59\% & RED \\
Desktop predictor $\rightarrow$ Orin PyTorch & 0.001632 & 0.000976 & 0.010471 & 98.69\% & RED \\
Orin PyTorch $\rightarrow$ TensorRT FP16 & 0.000058 & 0.000296 & 0.001180 & 99.68\% & GREEN \\
\bottomrule
\end{tabular}%
}
\end{table*}

% 5.2
\begin{figure*}[!t]
  \centering
  \includegraphics[trim={0 0 0 3.65cm}, clip, width=0.9\textwidth]{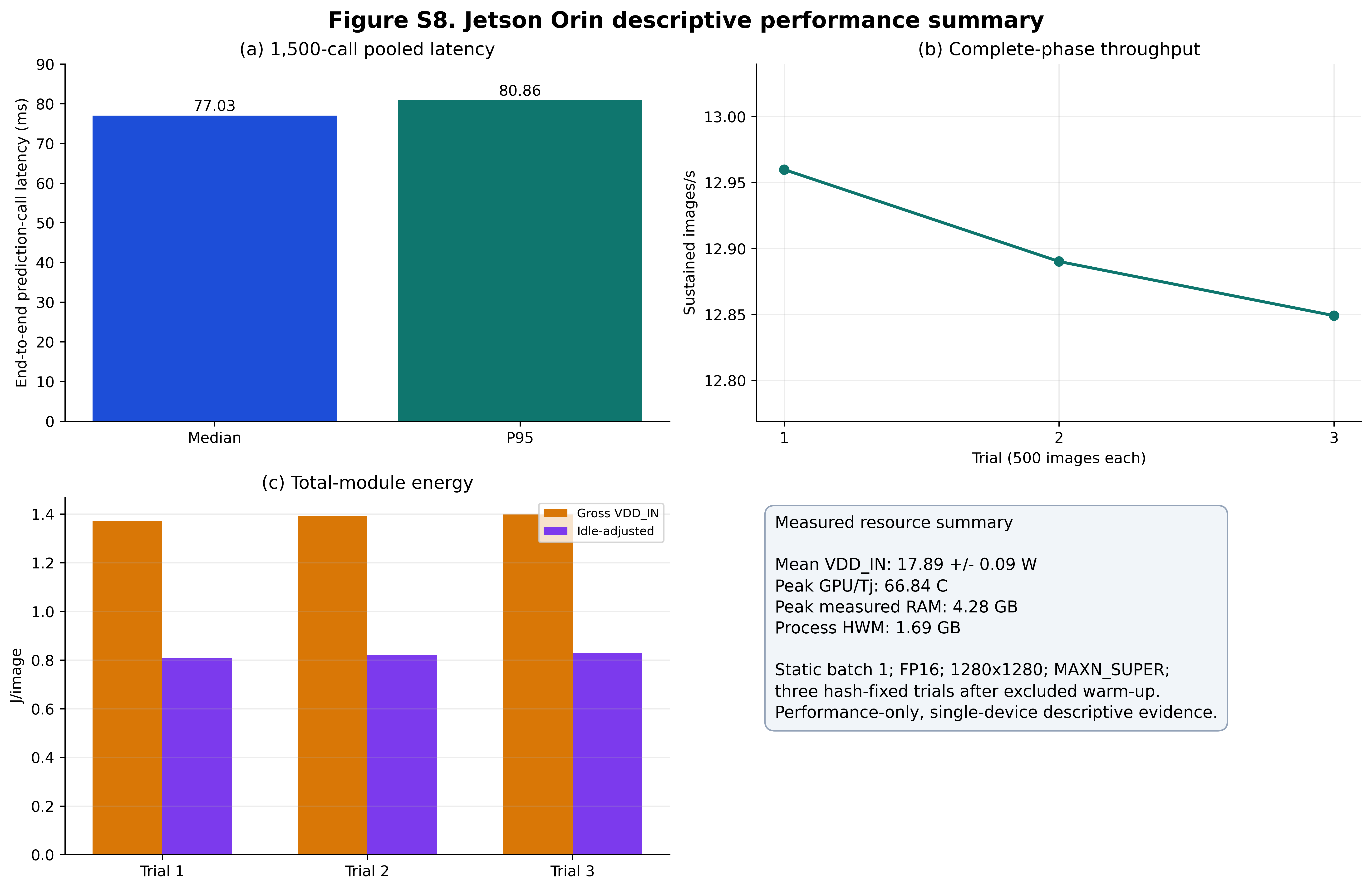}
  \vspace{-2ex}
  \caption{Per-trial performance on the Orin. The three trials repeat the same 500 measured development images, and energy refers to total module input. These timings exclude acquisition, network, and UI/display.}
  \label{Suppl_Fig_S9}
\end{figure*}

% 5.2
\begin{table}[H]
\centering
\caption{Orin descriptive measurements. Latency summarizes the 1,500 post-warm-up calls; $\pm$ denotes sample SD across the three trials for throughput, power, and energy. Fig.~\ref{Suppl_Fig_S9} shows their distributions.}
\small
\resizebox{\columnwidth}{!}{%
\begin{tabular}{ll}
\toprule
\textbf{Quantity} & \textbf{Details/Result} \\
\midrule
Prediction calls & 1,500 ($3\times500$) \\
End-to-end pred. latency, mean & 77.4499 ms \\
End-to-end pred. latency, median & 77.0253 ms \\
End-to-end pred. latency, P95 & 80.8553 ms \\
Inference latency, mean/med./P95 & 50.4095 / 49.8776 / 53.3658 ms \\
Preprocess latency, mean/med./P95 & 7.3512 / 7.3732 / 7.5158 ms \\
Postprocess latency, mean/med./P95 & 1.2382 / 1.2720 / 1.4024 ms \\
Sustained throughput & 12.8997{\scriptsize$\pm$0.0560} images/s \\
Mean total-module $\text{VDD\_IN}$ & 17.8941{\scriptsize$\pm$0.0944} W \\
Gross total-module energy & 1.38723{\scriptsize$\pm$0.01329} J/image \\
Engine-resident idle reference & 7.33604 W \\
Signed idle-adjusted energy & 0.81852{\scriptsize$\pm$0.01084} J/image \\
Peak GPU/Tj temperature & 66.843\textdegree C \\
Peak measured-phase RAM use & 4,280 MB \\
Maximum process RSS/HWM & 1,693,020,160 bytes (1.69 GB) \\
\bottomrule
\end{tabular}%
}
\end{table}

\section*{S5. Device Inference Agreement and Performance}
\subsection*{S5.1. Development-set inference agreement}
% Engineering agreement is decomposed rather than summarized as a single platform-equivalence claim.
For the within-Orin comparison, prediction-agreement F1 at IoU 0.90 is 0.9975 (5,961 matches; 5,977 PyTorch and 5,975 TensorRT predictions). The first two comparisons are RED, so agreement across the full desktop-to-device workflow is not established; the GREEN result applies only to the Orin PyTorch $\rightarrow$ TensorRT conversion.

\subsection*{S5.2. Orin Benchmarking}
The performance-only benchmark uses the checksum-bound STKD seed-1 static FP16 engine, input shape $1\times3\times1280\times1280$, batch 1, on an Orin Nano Super configuration with JetPack 7.2.1-b49 and TensorRT 10.16.2.10. Three trials repeat the same 500-image development sequence after 50 disjoint warm-ups per trial. Prediction calls are CUDA-synchronized; measured files are page-cached; telemetry is sampled every 100 ms under MAXN$_\text{SUPER}$ and requested fixed clocks.

VDD$_\text{IN}$ is total-module input power integrated over the complete measured phase; it is not GPU-only or model-only energy. Memory values are unified-memory and process observations, not a TensorRT engine-footprint estimate. The benchmark excludes acquisition, UI/display, network, cold start, concurrency, and complete-workflow latency. It does not repair the RED upstream agreement checks or establish clinical, deployment-equivalence, real-time-workflow, or production-readiness claims.

\end{document}